\documentclass[10pt,twocolumn,letterpaper]{article}

\usepackage{iccv}
\usepackage{times}
\usepackage{epsfig}
\usepackage{graphicx}
\usepackage{amsmath}
\usepackage{amssymb}

\usepackage{mathrsfs}

\usepackage[export]{adjustbox}

\usepackage{multirow}

\usepackage{bbding}
\usepackage{flushend}
\usepackage{lscape}
\usepackage{subcaption}
\usepackage[pagebackref=true,breaklinks=true,letterpaper=true,colorlinks,bookmarks=false]{hyperref}

\iccvfinalcopy % *** Uncomment this line for the final submission

\graphicspath{{fig/}}

\def\iccvPaperID{5378} % *** Enter the ICCV Paper ID here

\ificcvfinal\fi

\begin{document}

%%%%%%%%% TITLE
\title{Rethinking Zero-Shot Learning: A Conditional Visual Classification Perspective}

\author{Kai Li$^{1}$, Martin Renqiang Min$^{2}$, Yun Fu$^{1,3}$ \\
$^1$Department of Electrical and Computer Engineering, Northeastern University, Boston, USA \\
$^2$NEC Laboratories America \\
$^3$Khoury College of Computer Science, Northeastern University, Boston, USA \\
{\tt\small kaili@ece.neu.edu, renqiang@nec-labs.com, yunfu@ece.neu.edu}
% \and
% \and
}

\maketitle
\ificcvfinal\thispagestyle{empty}\fi

%%%%%%%%% ABSTRACT
\begin{abstract}
 Zero-shot learning (ZSL) aims to recognize instances of unseen classes solely based on the semantic descriptions of the classes. Existing algorithms usually formulate it as a semantic-visual correspondence problem, by learning mappings from one feature space to the other. Despite being reasonable, previous approaches essentially discard the highly precious discriminative power of visual features in an implicit way, and thus produce undesirable results.  We instead reformulate ZSL as a conditioned visual classification problem, i.e., classifying visual features based on the classifiers learned from the semantic descriptions. With this reformulation, we develop algorithms targeting various ZSL settings: For the conventional setting, we propose to train a deep neural network that directly generates visual feature classifiers from the semantic attributes with an episode-based training scheme; For the generalized setting, we concatenate the learned highly discriminative classifiers for seen classes and the generated classifiers for unseen classes to classify visual features of all classes; For the transductive setting, we exploit unlabeled data to effectively calibrate the classifier generator using a novel learning-without-forgetting self-training mechanism and guide the process by a robust generalized cross-entropy loss. Extensive experiments show that our proposed algorithms significantly outperform state-of-the-art methods by large margins on most benchmark datasets in all the ZSL settings. Our code is available at \url{https://github.com/kailigo/cvcZSL}
\end{abstract}

%%%%%%%%% BODY TEXT
\section{Introduction}
Deep learning methods have achieved revolutionary successes on many tasks in computer vision owing to the availability of abundant labeled training data \cite{zhang2019rnan,zhang2019mst,li2018support,li2019attnbn,li2019gain,li2019vsr}. 
However, labeling large-scale training data for each task is both labor-intensive and unscalable. Inspired by human's remarkable abilities to recognize instances of unseen classes solely based on class descriptions without seeing any visual example of such classes, researchers have extensively studied an image classification setting similar to the human learning called zero-shot learning (ZSL)~\cite{xian2018zero,schwartz2018delta,li2018discriminative,song2018selective}, in which labeled training images of seen classes and semantic descriptions of both seen classes and unseen classes are given and the task is to classify test images into seen and unseen classes.

Existing approaches usually formulate ZSL as a visual-semantic correspondence problem and learn the visual-semantic relationship from seen classes and apply it to unseen classes, considering that the seen and unseen classes are related in the semantic space~\cite{akata2015evaluation,zhang2017learning,kodirov2015unsupervised}.  These methods usually project either visual features or semantic features from one space to the other, or alternatively project both types of features to an intermediate embedding space. In the shared embedding space, the associations between the two types of features are utilized to guide the learning of the projection functions. 

However, these methods fail to recognize the tremendous efforts in obtaining these discriminative visual features over a large number of classes through training powerful deep neural network classifiers with a huge amount of computational and data resources, and thus essentially  discard  the  highly  precious  discriminative power  of visual features in an implicit way. In details, on one hand, the visual features used in most ZSL methods are extracted by some powerful deep neural networks (e.g., ResNet101) trained on large-scale datasets (e.g., ImageNet) \cite{xian2016latent}. These visual features are already highly discriminative; reprojecting them to any space shall impair the discriminability, especially to a lower dimensional space, because the dimension reduction usually significantly shrinks data variance. It is surprising that the majority of existing ZSL approaches try to transform the visual feature vectors in various ways \cite{li2018discriminative,song2018selective,kodirov2015unsupervised}. On the other hand, by nature of classification problems, the competition information among different classes are crucial for classification performance. But many ZSL approaches ignore the class separation information during training due to focusing on learning the associations between visual and semantic features, and fail to realize that ZSL is essentially a classification problem~\cite{zhang2017learning}.

Inspired by the above observations, we propose to solve ZSL in a novel conditional visual feature classification framework. In the proposed framework, we effectively generate visual feature classifiers from the semantic attributes, and thus intrinsically preserve the visual feature discriminability while exploiting the competing information among different classes. 
Within the novel framework, we propose various novel strategies to address different ZSL problems.

% various ZSL problems can be effectively addressed by our proposed novel strategies.  
% we approach various ZSL problems. 
% various ZSL In our new ZSL framework, we develop algorithms for various ZSL settings. 

For the conventional ZSL problem where only unseen classes are involved for evaluations, we propose to train a deep neural network that generates visual feature classifiers directly from the semantic attributes. We train the network with a Cosine similarity based cross-entropy loss, which mitigates the impact of variances of features from two different domains when calculating their correlations. Borrowing ideas from meta-learning, we train our model in an episode-based way by composing numerous ``fake'' new ZSL tasks, so that its 
generalizability to ``real'' new ZSL tasks during test is enhanced. 
For the generalized setting in which seen classes are included for ZSL evaluations, we concatenate the classifiers for seen classes and unseen classes to classify visual features for all classes. Since the classifiers for seen classes are trained with labeled samples, they are highly discriminative to discern whether an incoming image belongs to the seen classes or not. This desirable property prevents our method from significant performance drops when much more classes are involved for evaluations. For the transductive setting in which images of unseen classes are available during training \cite{song2018transductive}, we take advantage of these unlabeled data to calibrate our classifier generator using the pseudo labels generated by itself. To limit the harm of incorrect pseudo labels and avoid the model being over-adapted to new classes, we propose to use the generalized cross-entropy loss to guide the model calibration process under an effective learning-without-forgetting training scheme.

In summary, our contributions are as follows:
\begin{itemize}
\vspace{-3pt}
    \item We reformulate ZSL as a conditional visual classification problem, by which we can essentially benefit from high discriminability of visual features and inter-class competing information among training classes to solve ZSL problem in various settings.  
\vspace{-3pt}
    \item We propose various effective techniques to address different ZSL problems uniformly within the proposed framework.
    % in different settings. 
\vspace{-3pt}
    \item Experiments show that our algorithms significantly outperform state-of-the-art methods by large margins on most benchmark datasets in all the ZSL settings.
\end{itemize}

\section{Related Work}
% Hubness [36] states the phenomenon that the mapped semantic embeddings from images would be collapsed to hubs, which are near many other
% points without being similar to the class label in any meaningful way.

% ZSL emerges in situations where we have never seen a class before but get some semantic descriptions of the class. We need recognize the class based on the semantic descriptions. 
Zero-Shot Learning (ZSL) aims to recognize unseen classes based on their semantic associations with seen classes. 
The semantic associations could be within the 
human-annotated attributes \cite{song2018transductive,morgado2017semantically,annadani2018preserving},  word vectors \cite{frome2013devise,zhang2017learning,changpinyo2017predicting}, text descriptions \cite{lei2015predicting,yizhe_zsl_2017}, etc.
% The majority of the ex
In practice, ZSL is performed by firstly learning an embedding space where semantic vectors and visual features are interacted. Then, within the learned embedding space, the best match among 
semantic vectors of unseen classes is selected for the visual features of any given image of the unseen classes.

According to the embedding space used, existing methods can be generally categorized into the following three groups.
% those Some approaches select the semantic space as the embedding space and project visual features to the semantic space (\cite{lampert2014attribute,frome2013devise}
% Since the embedding space is often of high dimension, finding the best match of a given vector among many candidates shall inevitably encounter the hubness problem (\cite{radovanovic2010hubs}), $\textit{i.e.}$, some candidates 
% will be biased to be the best matches for many of the queries. 
% Depending on the chosen embedding space, the severeness of this problem varies. 
Some approaches select semantic space as embedding space and project visual features to semantic space \cite{lampert2014attribute,frome2013devise}. Projecting visual features into a often much lower-dimensional semantic space shall shrink the variance of the projected data points and thus aggravate the hubness problem, i.e., some candidates 
will be biased to be the best matches to many of the queries. 
% may aggravate the hubness problem.
Alternatively, some methods project both visual and semantic features into a common intermediate space \cite{akata2015evaluation,yang2018learning,zhang2015zero}. However, due to the lack of training samples from unseen classes, these methods are prone to classifying test samples into seen classes \cite{romera2015embarrassingly}. The third category of methods  choose the visual space as the embedding space and learn a mapping from the semantic space to visual space \cite{zhang2017learning}. Benefiting from the abundant data diversity in visual space, these methods can mitigate the hubness problem to some extent. 

Recently, a new branch of methods come out and approach ZSL in virtue of data augmentation, either by variational auto-encoder (VAE) 
\cite{mishra2018generative} or Generative Adversarial Network (GAN) \cite{chen2018zero,xian2018feature,felix2018multi,zhu2018generative,yizhe_abp_2019}. These methods learn from visual and semantic features of seen classes and produce generators that can generate synthesized visual features based on class semantic descriptions.  Then, synthesized visual features are used to train a standard classifier for object recognition. 

% Our method avoids this problem.

% By nature of a classification problem, both intra-class compactness (visual features of the same classes are assigned with the same label) and inter-class separability (visual features of different classes are assigned with different labels) are exploited, hence resulting in a better mapping.

ZSL may turn easier when unlabelled test samples are available during training, i.e., the so-called transductive ZSL. This is because unlabelled test samples can be utilized to help reach clearer decision boundaries for both seen and unseen classes.  In fact, it is more like a semi-supervised learning problem. Propagated Semantic Transfer (PST) \cite{rohrbach2013transfer} conducts label propagation from seen classes to unseen classes through exploiting the class manifold structure. Unsupervised Domain Adaption (UDA) \cite{kodirov2015unsupervised} formulates the problem as a cross-domain data association problem and solves it by regularized sparse coding. Quasi-Fully Supervised Learning (QFSL) \cite{song2018transductive} aims to strength the mapping from visual space to semantic space by explicitly requiring
the visual features being mapped to the categories (seen and unseen) they belong.  

% the labeled source images are mapped to several fixed points specified by the source categories, and the unlabeled target images are forced to be mapped to other points specified by the target categories.

% Our method solves the 

Unlike the above methods, we approach ZSL from the perspective of 
conditioned visual feature classification. Perhaps most similar to our algorithms are \cite{lei2015predicting,wang2018zero}, which approach ZSL also by generating classifiers. However, \cite{lei2015predicting} projects visual features to a lower dimensional space, harming discriminability of the visual features. \cite{wang2018zero} uses graph convolutional network to model the semantic relationships and output classifiers. However, it requires  categorical relationship as additional inputs. We instead generate classifiers directly from attributes by a deep neural network and train the model with a novel cosine similarity based cross-entropy loss. 
Besides, neither of the two methods uses episode-based training to enhance model adaptability to novel classes. Moreover, they are only feasible for the conventional ZSL setting, while our method is flexible for various ZSL settings.

% requires categorical relationship
% However, our algorithms differ with them 

% neural network generation, following \cite{lei2015predicting,wang2018zero}. We formulate bridging the semantic space and the visual space as a visual feature classification problem conditioned on the semantic features.  
% Unlike \cite{wang2018zero} which predicts visual classifier for each visual category using Graph Convolutional Network and takes as input both semantic embedding and categorical relationship, we learn a deep neural network to generate classifier directly from the semantic embedding.
% Our method is based on \cite{lei2015predicting}, and we revisit the used techniques and reveal that they can used in more effective way. Meanwhile, our method can be flexibly extended to the transductive setting and reach better performance by taking advantage of the unlabelled test samples. 

% in a different way from the above methods. 
% We approach ZSL
% \cite{wang2018zero}

% Unlike the above methods, following \cite{lei2015predicting}, we

% we formulate bridging the semantic space and the visual space
% as a visual feature classification problem conditioned on the semantic features.  
% We learn a deep neural network that generates classifier weights for the visual features when fed with the corresponding semantic features. 

\section{Method}
Zero-shot learning (ZSL) is to recognize objects of unseen classes given only semantic descriptions of the classes.  Formally, suppose we have three sets of data $\mathcal{D}=\{\mathcal{D}_s, \mathcal{D}_a, \mathcal{D}_u\}$, where $\mathcal{D}_s=\{\mathcal{X}_s, \mathcal{Y}_s\}$ and $\mathcal{D}_u=\{\mathcal{X}_u, \mathcal{Y}_u\}$ are training and test sets, respectively. 
% the test set, with 
$\mathcal{X}_s$ and $\mathcal{X}_u$ are the images, while $\mathcal{Y}_s$ and $\mathcal{Y}_u$ the corresponding labels.  There is no overlap between training classes and test classes, i.e., $\mathcal{Y}_s \cap \mathcal{Y}_u=\emptyset$.  The goal of ZSL is to learn transferable information from $\mathcal{D}_s$ that can be used to classify unseen classes from $\mathcal{D}_u$, with the help of semantic descriptions $\mathcal{D}_a=\mathcal{A}_s\cup\mathcal{A}_u$ for both seen ($\mathcal{A}_s$) and unseen ($\mathcal{A}_u$) classes. $\mathcal{D}_a$ can be human-annotated class attributes \cite{xian2018feature} or articles describing the classes \cite{yizhe_zsl_2018}. 

We solve ZSL in a conditional visual feature classification framework. Specifically, we predict the possibility $p(y|\textbf{x}; \textbf{a}_y)$ of an image $\textbf{x}$ belonging to class $y$ given the semantic description $\textbf{a}_y$ of the class, where $y\in\mathcal{Y}_u$ in the standard setting, while $y\in\mathcal{Y}_s\cup\mathcal{Y}_u$ in the generalized setting. When $\mathcal{X}_u$ is available during training, we call the problem transductive ZSL. For convenience, sometimes we call the setting inductive ZSL where $\mathcal{X}_u$ is unavailable.  

% We will first introduce our algorithm for the conventional ZSL setting and then show how we extend it to deal with the other two settings.

% As a comparison, when $\mathcal{X}_u$
% For standard ZSL setting, $\textbf{x}\in$
% We cope this problem by revisiting the neural network generation technique and revealing that it can be utilized in a more effective way for both inductive setting ($\mathcal{X}_u$ is unavailable during training) and transductive setting ($\mathcal{X}_u$ is available during training).
% \begin{figure}
%   \begin{center}
%     \includegraphics[width=1.0\linewidth]{./figs/framework.pdf}         
%   \end{center}
%   \vspace{-5pt}
% \caption{Framework of the proposed method.}
% \label{fremework}   
% \vspace{-10pt}
% \end{figure}
% Our training procedure is based on 

\subsection{Zero-Shot Learning}
By approaching ZSL in virtue of visual classification conditioned on attributes, we need to generate visual feature classifiers from the attributes. We achieve this by learning a deep neural network $f$ which takes a semantic feature vector of a class as input and outputs the classifier weight vector for the class.
% By the simple machine learning principle that test and train conditions must match, we need to mimic 
% $f$
% the same way it is going to be test. 
Since the model $f$ is going to generate classifiers for novel classes when tested, we adopt the episode-based training mechanism, an effective and popular technique in meta-learning~\cite{vinyals2016matching,finn2017model,li2019on}, to mimic this scenario during training. 
% To our best knowledge, our work here is the first to introduce episode-based training into ZSL.

% Inspired by recent progress in meta-learning \cite{vinyals2016matching,finn2017model}, for the first time as to our best knowledge, we introduce episode-based training for ZSL.

% We constantly sampling new ZSL tasks from minibatch to minibatch. 
The key to episode-based training is to sample in each mini-batch a ``fake'' new task that matches the scenario where the model is tested. This process is called an episode. The goal is to expose the model with numerous ``fake'' new tasks during training, such that it can generalize better for real new tasks when tested. To construct a ZSL episode,
% called 
% as an episode.
% sampling new ZSL tasks
% This training procedure matches exactly how it will be tested when presented with the semantic descriptions of new classes and output a classifier to classify visual features of the classes. 
% More specifically, to train $f$, 
we randomly sample from $\mathcal{D}_t=\{\mathcal{X}_t, \mathcal{Y}_t\}$ and $\mathcal{A}_t$ a ZSL task 
$\mathcal{T}=\{\mathcal{V}, \mathcal{A}\}$ 
where 
% $\mathcal{V}$ 
$\mathcal{V}=\{\textbf{x}_{i, j}\}^{N}_{i=1}, y_j\}^{M}_{j=1}$
% $\mathcal{V} = \{(\textbf{x}_1, y_1 ), (\textbf{x}_2 , y_2 ), \cdots, (\textbf{x}_{M\times N} , y_{M\times N})\}$ 
contains samples for $M$ classes, $N$ samples per classes. Note for each sample $(\textbf{x}_{i, j}, y_j)$,
we dismiss its global (dataset-wise) label and replace it with a local (minibatch-wise) label (i.e., $y_j\in\{1, 2, \cdots, M\}$), while still maintaining the class separation (samples of the same global label still with the same local label). This is to cut off the connections across tasks induced by the shared global label pool, so that each mini-batch is treated as a new task.
$\mathcal{A}= \{\textbf{a}_1, \textbf{a}_2, \cdots, \textbf{a}_M\}$ is the associated $M$ attribute vectors. 
For each task $\mathcal{T}=\{\mathcal{V}, \mathcal{A}\}$, $f$ generates a classifier for the $M$ sampled classes as
\begin{equation}
\mathbf{W} =  f(\mathcal{A}).
\label{weight_generation} \\
\end{equation}
% where $\textbf{A}=(\textbf{a}_1, \textbf{a}_2, \cdots, \textbf{a}_M)$ is the attribute matrix of the $M$ classes. 
% $\mathbf{W} \in \mathbb{R}^{d_w\times M}$ is the classifier weight matrix. 
With the classifier $\mathbf{W}$, we can calculate classification scores of visual features from $\mathcal{V}$. Rather than using the extensively used dot product, we use cosine similarity.  
% of a standard Logistic regression classifier. 
% ($\textbf{a}_i\in\mathcal{A}, \forall i$) of $N$ classes,  
% we generate the classifier weights for the $N$ classes as
% where $f$ is a neural network that maps semantic descriptions to classifier weights. 

% Using $\mathbf{W} \in \mathbb{R}^{d_w\times N}$ as the classifier, we calculate t

% will then be used to calculate classification score with the feature embedding $\textbf{e}_x=f_{\theta}(\textbf{x})\in \mathbb{R}^{d}$ of any given image \textbf{x} belonging to the $N$ classes. $f_{\theta}$ is the feature embedding network. It is noted that we generate classifier weights for the visual embedding space. 

% This is different from \cite{lei2015predicting} where classifier weights are generated for another lower dimensional space to which the visual feature embedding $\textbf{e}_x$ is reprojected. 
% We choose the visual embedding space as the classification space because it can preserve the discriminability of feature embedding, which is often extracted by some powerful extraction feature network. Re-projecting $\textbf{e}_x$ to any other space often harms the discriminability. Meanwhile, re-projecting visual embedding to lower dimensional space will shrink the visual variance, which aggravates the notorious hubness problem.
\begin{table}[t]
  \small
  \centering
  \begin{tabular}{l}
    \hline
    \noindent \textbf{Algorithm 1.} Proposed ZSL approach \\\hline 
    \textbf{Input:} Training set $\mathcal{D}_s=\{\mathcal{X}_s, \mathcal{Y}_s\}$ and attributes $\mathcal{A}_s$. \\ 
    \textbf{Output:} Classifier weight generation network $f$ \\ \hline
    \textbf{while} not done \textbf{do}\\
    \hspace{3mm} 1. Randomly sample from $\mathcal{D}_s$ and $\mathcal{A}_s$ a ZSL task \\ 
    % \hspace{6mm} ,   \\ 
    \hspace{6.2mm} $\mathcal{T}=\{\mathcal{V}, \mathcal{A}\}$, where $\mathcal{V}=\{\{\textbf{x}_{i, j}\}^{N}_{i=1}, y_j\}^{M}_{j=1}$ \\
    \hspace{6.2mm} and $\mathcal{A}=\{\textbf{a}_j\}^{M}_{j=1}$. \\
    \hspace{3mm} 2. Calculate loss according to Eq. \eqref{obj1} \\ 
    \hspace{3mm} 3. Update $f$ through back-propagation. \\
    \textbf{end while} \\
    \hline
  \end{tabular}
  \vspace{-5pt}
\end{table}

% We adopt the cosine similarity based cross entropy loss to train the weight generator $g_\phi$.
% the cosine similarity based softmax loss function.
% based classifier

\noindent \textbf{Cosine similarity based classification score function}. 
Traditional multi-layer neural networks use dot product between the output vector of previous layer and the incoming weight vector as the input to activation function. \cite{luo2017cosine,gidaris2018dynamic} recently showed that replacing the dot product with cosine similarity can bound and reduce the variance of the neurons and thus result in models of better generalization. 
Considering that 
% The cosine similarity based score function shall suit our problem well because
we are trying to calculate the correlation between data from two dramatically different domains, especially for the attribute domain in which the features are discontinuous and have high variances. 
Using cosine similarity shall mitigate the harmful effect of the high variances and bring us desirable Softmax activations. With this consideration, we
define our classification score function as 
% limit and bound the variance and bring us desirable Softmax activations.
% Inspired by this, 
% rather than using dot product to calculate classification score as in \cite{lei2015predicting}, 
% we define our score function as 

% we define our cosine similarity 
% \cite{gidaris2018dynamic} further showed that using the cosine similarity instead of dot product for calculating classification score in the last fully-connected layer of deep neural network brings benefit for classification. 
% using the softmax function:
% \mathbf{e}_x & = & f_\theta (\mathbf{x}),\\
% \begin{eqnarray}
% p(y=n|\textbf{e}_x) & = & \frac{\exp(s\cos(\mathbf{w}_y, \mathbf{e}_x))}{\sum_{j=1}^N \exp(s\cos(\mathbf{w}_j, \mathbf{e}_x))},
% \label{probability} \\
% \mathbf{w}_j & = & g_\phi(\textbf{a}_j),
% \end{eqnarray}
% where $\eta$
% We adopt this technique to train our weight generator $g_\phi$.  The classification score of a sample ($\textbf{e}_x$, $y$) is calculated as
% where $\mathbf{w}_j$ is the classification weight vector for class $j$, which is generated by neural network $g_{\phi}$ 
% semantic description vector of the class as input, 
\begin{equation}
p(y=i|\textbf{x}) = \frac{\exp(\sigma\cos(\mathbf{w}_i, \mathbf{x}))}{\sum_{j=1}^N \exp(\sigma\cos(\mathbf{w}_j, \mathbf{x}))},
\label{probability} 
\end{equation}
where $\sigma$ is a learnable scalar controlling the peakiness of the probability distribution generated by the Softmax operator.
$\mathbf{w}_i$ is the classifier weight vector for class $i$. 

% Compared with dot product, Cosine similarity
% (visual and attribute) where data
% Considering that 
% Compared with dot product, Cosine similarity can limit and bound the input variance, bringing us desirable Softmax activations. 

% generating from the semantic description of the class using Eq. \eqref{weight_generation}. 
% $\textbf{e}_x$ is the feature embedding of an image $\textbf{x}$ belong belong to class $y$ using a feature embedding neural network $f_\theta$:
% \begin{equation}
% \mathbf{e}_x  =  f_\theta(\textbf{x}),
% \label{feature_embedding} 
% \end{equation}
% $\mathbf{x}$ is the input image, $\mathbf{a}_j$ is the attribute vector for class $j$ for ZSL, $\mathbf{x}_{i,j}$ is the $i$-th input image of class $j$ for FSL, $j = 1, ..., N_f$, and $N_f$ is the number of shots for FSL. 
With this definition, the loss of a typical ZSL task $\mathcal{T}$ 
% based on a cross-entropy loss function 
is defined as follows,
% $\mathbf{w}_{(\cdot)}=f_{\theta}(\cdot)$ is the classifier weights. 
% \begin{equation}
% \mathcal{L}(\theta, \phi) = \sum_{(\textbf{x},y)\in\mathscr{T}}\big[-s\cos(\mathbf{w}_y, 
% f_\theta(\mathbf{x})) + \log(\sum_{j=1}^N\exp(s\cos(\mathbf{w}_j, f_\theta(\mathbf{x}))))\big] + \lambda\|\phi\|_2,
% \label{loss_function}
% \end{equation}
\begin{equation}
\begin{array}{cl}
\mathcal{L} & =  \sum_{(\textbf{x},y)\in\mathcal{T}}\big[-\sigma\cos(\mathbf{w}_i, 
\mathbf{x}) + \\
& \log(\sum_{j=1}^N\exp(\sigma\cos(\mathbf{w}_j, \mathbf{x})))\big] + \lambda\|\phi\|_2,
\end{array}
\label{obj1}
\end{equation}
where $\lambda$ is a hyper-parameter weighting the $l_2$-norm regularization of the learnable parameters of neural network $f_{\phi}$.

\textbf{Algorithm 1} outlines our training procedures.

\subsection{Generalized Zero-Shot Learning}
With the learned classifier generator $f$, given attributes of unseen classes $\mathcal{A}_u$ in the test stage, we generate the corresponding classifier weights $\mathbf{W}_u=f(\mathcal{A}_u)$ and use them to classify visual features of unseen classes $\mathcal{X}_u$ according to Eq. \eqref{probability}.
% where attributes of both  $\mathcal{D}_a=\mathcal{A}_s\cup\mathcal{A}_u$ are candidate match for any give image, 

When both seen and unseen classes are involved for evaluations, i.e., the generalized ZSL setting, we combine the classifiers for both seen and unseen classes to classify images from all classes. Specifically, with $\mathcal{A}_u$ and $\mathcal{A}_s$, we can get classifiers 
$\mathbf{W}_u=f(\mathcal{A}_u)$ and $\mathbf{W}_s=f(\mathcal{A}_s)$ for unseen and seen classes, respectively. 
We use their concatenation $\textbf{W}_b=[\mathbf{W}_u, \mathbf{W}_s]$ as the classifier for all classes. 
% for unseen classes and  for seen classes, 
% outputing . We use $\textbf{W}_b$ to classify features from both seen and unseen classes. 
% we concatenate the classifier weights 

It is worth noting that since $f$ has already been trained with labeled samples, the resulting $\mathbf{W}_s$ should be very discriminative to discern whether an incoming image belongs to the seen classes or not. As will be shown later in the experiments, 
this desirable property prevents our method from significant recognition accuracy drops when much more classes are involved for evaluations.

% We concate
% or given attributes $\mathcal{S}=\mathcal{A}_t\cup\mathcal{A}_u$ for all (seen and unseen) classes in generalized ZSL setting, we generate classifier weights $\mathbf{W}_b$ for all classes as
% \begin{equation}
% \mathbf{W}_b=g_{\phi}(\mathcal{S}).
% \label{weight_test_all}
% \end{equation}
% With $\mathbf{W}_u$ or $\mathbf{W}_b$, we can calculate the classification scores for images from unseen or seen classes according to the Eq. (2) in the main article.

\subsection{Transductive Zero-Shot Learning}
\begin{figure}
  \begin{center}
    \includegraphics[width=0.9\linewidth]{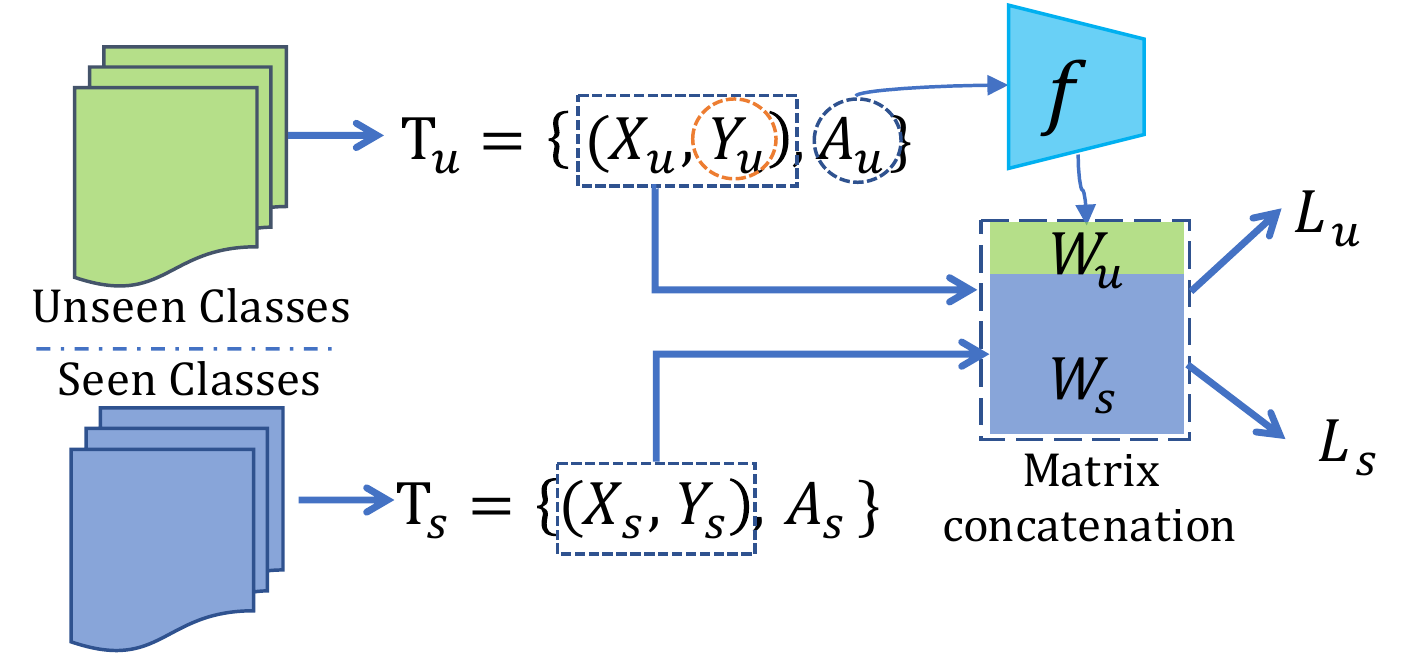}     
  \end{center}
  \vspace{-10pt} 
\caption{Illustration of the transductive ZSL algorithm. We sample ZSL tasks $\mathcal{T}_s$ from seen classes and $\mathcal{T}_u$ from unseen classes (with pseudo labels). 
% from both unseen classes ($\mathcal{T}_u$) and seen classes ($\mathcal{T}_s$). 
The classifier $\textbf{W}_u$ generated from $\textbf{A}_u$ are concatenated with classifier $\textbf{W}_s$ to classify visual features from both $\mathcal{T}_u$ and $\mathcal{T}_s$, which results in loss $\mathcal{L}_u$ and $\mathcal{L}_s$, respectively. The pseudo labels for unseen classes are updated in a self-training way.
}
% During training, we jointly update $f$ and $\textbf{W}_s$ using $$
\label{trans}   
\vspace{-5pt} 
\end{figure}

Thanks to the conditional visual classification formulation of ZSL, the above inductive approach can be readily adapted to the transductive ZSL setting. We can utilize test data during training to calibrate our classifier generator and output classifiers of better decision boundaries for both seen and unseen classes. We achieve this in virtue of self-training. Specifically, we alternate between generating pseudo labels for images of unseen classes using the classifier generator and updating it using the generated pseudo labels. With this idea, two key problems need to be solved. The first is how to prevent the generator from over-adapting to unseen classes such that the knowledge previously learned from seen classes is lost, resulting in unsatisfactory performance for seen classes. The second is how to avoid the generator being impaired by the incorrect pseudo labels. We propose a novel self-training based transductive ZSL algorithm to avoid both problems. Figure \ref{trans} illustrates our algorithm.

% avoid both two problems by introducing generalized cross-entropy loss under a novel training mechanism that can maintain knowledge previously learned when handling new tasks. 
To generate pseudo labels for test images $\mathcal{X}_u$, we first generate classifier weights $\textbf{W}_u$ for unseen classes as
\begin{equation}
\mathbf{W}_u=f(\mathcal{A}_u).
\label{weight_unseen}
\end{equation}
With $\mathbf{W}_u$, we calculate classification score $\textbf{S}$ of $\mathcal{X}_u$ according to Eq. $\eqref{probability}$. 
% Based on $\textbf{S}$, we can assign 
Pseudo labels $\tilde{\mathcal{Y}}_u$ of $\mathcal{X}_u$ can be obtained from $\textbf{S}$. There inevitably exist noises among $\tilde{\mathcal{Y}}_u$. 
We propose to mitigate their impact by a novel classification score peakiness based filtering strategy. 
% There inevitably are 
% incorrect
% pseudo labels within $\tilde{\mathcal{Y}}_u$. So, before using the pseudo labels for calibrating $f$, 
%  to limit the noise level.

% We propose a novel strategy to filter $\tilde{\mathcal{Y}}_u$ according to the peakiness of the classification scores for label assignment.
Let $\textbf{s}^i \in \mathbb{R}^{N_u}$ be the classification score of $\textbf{u}_i\in\mathcal{X}_u$ according to all the $N_u$ classes. 
Let $s^i_{y_m}$ and $s^i_{y_n}$ be the highest and second highest score among $\textbf{s}^i$. The pseudo label assigned to $\textbf{u}_i$ should be ${y_m}$. However, we regard this assignment as a ``confident'' one unless $s^i_{y_m}$ is peaky enough: 
\begin{equation}
\frac{s^i_{y_m}}{s^i_{y_n}} > \gamma,
\label{peakiness}
\end{equation}
where $\gamma$ is a threshold controlling the peakiness. This constraint prevents ambiguous label assignment from being exploited for classifier generator calibration. 

% Please note that correct label assignments shall have high peakiness. 
% As $f$ being progressively calibrated towards unseen classes, more samples will meet this peakiness condition, which in return 
% So, the number of confident assignments will be progressively enlarged with the refinement of $g_{\phi}$ towards unseen classes. 
% The larger collection of test images with high confidence (accuracy) 
% leads to better calibration of $f$.

After obtaining the confident set $\hat{\mathcal{D}}_u=\{\hat{\mathcal{X}}_u, \hat{\mathcal{Y}}_u\}$, as well as the the corresponding attributes $\hat{\mathcal{A}}_u$, we can use them to adjust $f$. However, finetuning $f$ with only $\hat{\mathcal{D}}_u$ shall cause strong bias towards unseen classes such that the knowledge previously acquired about seen classes will be forgotten after a few iterations. What is worse, the incorrect pseudo labels among $\hat{\mathcal{Y}}_s$ may damage $f$ when they are of a high portion. We propose a novel learning-without-forgetting training scheme to avoid this.

Along with sampling a ZSL task $\mathcal{T}_u$ from ($\hat{\mathcal{D}}_u$, $\hat{\mathcal{A}}_u$) to calibrate $f$ to unseen classes,
% $\hat{\mathcal{D}}_u=\{\hat{\mathcal{X}}_u, \hat{\mathcal{Y}}_u\}$ and $\hat{\mathcal{A}}_u$ to 
we sample another ZSL task $\mathcal{T}_s$ from ($\mathcal{D}_s$, $\mathcal{A}_s$) to keep the memory of $f$ to seen classes and dilute the impact of noisy labels from $\mathcal{T}_u$. Further, while updating $f$, we update as well classifier $\mathbf{W}_s$ to adjust the decision boundaries of seen classes towards unseen ones.
% and refine it 
% $\mathbf{W}_s$ is initialized by 
% \begin{equation}
% \mathbf{W}_s=f(\mathcal{A}_s)
% % \label{obj_gzsl}
% \end{equation}
% using attributes of all seen classes $\mathcal{A}_s$ prior to the transductive training stage. It preserves the discriminative information learned using only seen classes. 

% During the transductive training stage, combined with classifier weights generated by $f$ using attributes of novel classes, $\mathbf{W}_s$ is trained to handle well both tasks $\mathcal{T}_u$ and $\mathcal{T}_s$. It should noted that while $f$ is in a relatively higher risk of being over-adapted to unseen classes and impaired by incorrect labels, $\mathbf{W}_s$ is less impacted because it does not handle unseen data directly.

% $f$ is trained to handle both tasks well in each time. 
% The noise-free data from $\mathcal{T}_s$ on the one hand dilute the impact of noisy labels from $\mathcal{T}_u$, while on the hand keep the memory of $f$ to seen classes.
% To avoid severe forgetting of seen classes while adapting the unseen classes and also limit the impact of incorrect pseudo labels, we propose the following unique training mechanism.

\begin{table}[t]
  \small
  \centering
  \begin{tabular}{l}\hline
    \noindent \textbf{Algorithm 2.} Proposed approach for transductive ZSL \\\hline 
    \textbf{Input:} Training set $\mathcal{D}_s=\{\mathcal{X}_s, \mathcal{Y}_s\}$, attribute set \\     
    \hspace{2mm} $\mathcal{D}_a=\mathcal{A}_s\cup\mathcal{A}_u$, and test images $\mathcal{X}_u$, parameters $\gamma$ and $q$ \\         
    \textbf{Output:} Class label $\tilde{\mathcal{Y}}_u$ of $\mathcal{X}_u$, weight generator $f$, \\
    \hspace{2mm} classifier weight $\textbf{W}_s$ for seen classes.  \\ \hline            
    1. Obtain $f$ with $\mathcal{D}_s$ and $\mathcal{A}_s$ using \textbf{Algorithm 1}. \\ 
    2. Obtain $\textbf{W}_s=f(\mathcal{A}_s)$. \\        
    \textbf{for} $r =1, 2, ... N_r$ \textbf{do} \\
    \hspace{3mm} 3. Calculate classifier weights for unseen classes\\ 
    \hspace{6mm}    $\textbf{W}_u=f(\mathcal{A}_u)$. \\
    \hspace{3mm} 4. Generate pseudo labels $\tilde{\mathcal{Y}}_u$ for $\mathcal{X}_u$ according to Eq. \eqref{probability}. \\    
    \hspace{3mm} 5. Select confident test set $\hat{\mathcal{D}}_u=\{\hat{\mathcal{X}}_u, \hat{\mathcal{Y}}_u\}$ and $\hat{\mathcal{A}}_u$ \\
    \hspace{6mm}    based on Eq. \eqref{peakiness}. \\
    % \hspace{3mm}    \textbf{while} not done \textbf{do} \\
    \hspace{3mm}    \textbf{for} $i =1, 2, ... N_i$ \textbf{do} \\
    \hspace{6mm} 6. Sample ZSL tasks $\mathcal{T}_s$ from ($\mathcal{D}_s$, $\mathcal{A}_s$), and \\
    \hspace{9mm} $\mathcal{T}_u$ from ($\hat{\mathcal{D}}_u$ $\hat{\mathcal{A}}_u$).\\
    \hspace{6mm} 7. Calculate loss according to Eq. \eqref{obj_gzsl}. \\         
    \hspace{6mm} 8. Update $f$ and $\textbf{W}_s$ through back-propagation. \\         
    \hspace{3mm} \textbf{end while} \\
    \textbf{end while} \\
    \hline
  \end{tabular}
  \vspace{-5pt}
\end{table}

Moreover, we introduce the very recently proposed generalized cross-entropy loss \cite{zhang2018generalized} to handle task $\mathcal{T}_u$ and limit the impact of incorrect pseudo labels to the classifier weight generator:
\begin{equation}
\mathcal{L}_u = \sum_{(\textbf{x}_u, y_u)\in\mathcal{T}_u} \frac{1-(\textbf{w}_{y_u})^q}{q},
\label{loss_q}
\end{equation}
where $\textbf{w}_{y_u}$ is the possibility of $\textbf{x}_u$ belonging to class ${y_u}$, which is calculated according to Eq. \eqref{probability}. $q\in$ (0,1] is a hyper-parameter of which a higher value is preferred when the noise level is high. It can be shown that Eq. \eqref{loss_q} turns to Eq. \eqref{obj1} when $q$ infinitely approaches 0. On the other hand, it turns to the Mean Absolute Error (MAE) loss when $q=1$. Cross-entropy loss is powerful for classification tasks but noise-sensitive, while MAE loss performs worse for conventional classification task but is robust to noisy labels. Tuning $q$ between 0 and 1 fits different noise levels. 

By handling $\mathcal{T}_u$ with generalized cross-entropy loss and $\mathcal{T}_s$ with conventional cross-entropy loss, our loss function for the transductive ZSL is as follows:
% The loss function for a typical transductive ZSL task can be calculated as
\begin{equation}
\mathcal{L}(\phi, \textbf{W}_s) = \mathcal{L}_u + \mathcal{L}_s,
\label{obj_gzsl}
\end{equation}
where $\mathcal{L}_s$ is defined in Eq \eqref{obj1}. \textbf{Algorithm 2} outlines the training procedures.

\section{Experiments}

\subsection{Datasets and Evaluation Settings} 
We employ the most widely-used zero-shot learning datasets for performance evaluation, namely, 
CUB \cite{welinder2010caltech}
AwA1 \cite{lampert2014attribute},
AwA2 \cite{xian2018zero},
% CUB \cite{wah2011multiclass},
SUN \cite{patterson2012sun} and aPY \cite{farhadi2009describing}. The statistics of the datasets are shown in Table \ref{Table_stat}. We follow the GBU setting proposed in \cite{xian2018zero} and evaluate both the conventional ZSL setting and the generalized ZSL (GZSL) setting. In the conventional ZSL, test samples are restricted to the unseen classes, while in GZSL, they may come from either seen classes or unseen classes. For both settings, we use top-1 (T1) Mean Class Accuracy (MCA) as the evaluation metric in our experiments. For GZSL, we evaluate the MCA for both seen ($S$) and unseen classes ($U$), and also calculate their harmonic mean $H=2*U*S/(U+S)$.

% Following the previous methods, we use Mean Class Accuracy (MCA) as the evaluation metric in our experiments, which measure mean
% To compare the performances, we adopt the Mean Class Accuracy (MCA) as the evaluation metric in our experiments:

% \begin{equation}
% MCA = \frac{1}{|Y|} \sum_{y\in Y} acc_y, 
% \label{mca}
% \end{equation}

% where $\textit{acc}_y$ is the top-1 accuracy on the test data from class $y$. In the conventional settings, MCA on only the target test data ($MCA_t$) is considered ($Y=Y_t$ in Eqn. 7). In the generalized settings, the search space at evaluation time is not restricted to the target classes, instead the the source classes are also included. Meanwhile, the test instances come from not only the target dataset, but also the source dataset ($Y = Y_s + Y_t$ in Eq. \label{mca}). Therefore, we adopt MCAt, MCAs (MCA on the source test data) and their harmonic
% mean (H) as the evaluation metrics.
% \begin{equation}
% H = \frac{2 * MCA_s * MCA_t}{MCA_s + MCA_t}
% \end{equation}
\begin{table}
  \footnotesize
  \renewcommand{\tabcolsep}{5.0pt}
  \begin{center}
    \begin{tabular}{c|c|cccccc}\hline   
     \multicolumn{2}{c|}{}                & CUB    & AwA1    & AwA2     & aPY   & SUN \\ \hline   
     \multirow{2}{*}{\#Class}  & \#Seen     & 150  & 40     & 40        & 20    & 645 \\
                        & \#Unseen     & 50 & 10     & 10     & 12       & 72  \\ \hline
     \multicolumn{2}{c|}{\# VisDim} &2048     & 2048   & 2048  & 2048   & 2048 \\ \hline
     \multicolumn{2}{c|}{\# AttDim} & 312  & 85     & 85 & 312   & 102 \\ \hline
    \end{tabular}
  \end{center} 
  \vspace{-5pt} 
  \caption{Information of zero-shot classification datasets.}
  \label{Table_stat}
  \vspace{-5pt}
\end{table}

\subsection{Implementation details}
Following \cite{xian2018zero}, we use ResNet101 \cite{he2016deep} trained on ImageNet for feature extraction,
which results in a 2048-dimension vector for each input image. 
The classifier generation model $f$ consists of two pairs of FC+ReLU layers, i.e., FC-ReLU-FC-ReLU, which 
maps semantic vectors to visual classifier weights. The dimension of the intermediate hidden layer is 1600 for all the five datasets. We train $f$ with Adam optimizer and a learning rate $10^{-5}$ for all datasets by 1,000,000 randomly sample ZSL tasks. Each task consists of 32 randomly sampled classes, 4 samples for each class, i.e., $M=32$ and $N=4$, except aPY where we set $M=16$ and $N=4$ because there are in total only 20 classes for training. The hyper-parameter $\lambda$ is chosen as $10^{-4}$, $10^{-3}$,  $10^{-3}$, $10^{-5}$ and $10^{-4}$ for AwA1, AwA2, CUB, SUN and aPY, respectively. 

For transductive ZSL, the experimental setting is the same as that in the corresponding inductive case for each dataset. For all the datasets, we update the pseudo labels of unseen classes every 10,000 iterations and execute 50 updates, i.e., $N_r=50$ and  $N_i=10,000$. We apply $\gamma=1.2$ and $q=0.5$ for all the datasets.  We develop our algorithms based on PyTorch.

\subsection{Ablation Studies} 
% For the inductive ZSL setting, our method differ from \cite{lei2015predicting} 
% By 
By formulating ZSL as a visual classification problem conditioned on the attributes, we can naturally benefit from the high discriminability of visual features. Meanwhile, to combat with the significant variance of visual and attribute features, we propose to replace the widely-used dot product with cosine similarity to calculate the classification score. Moreover, we introduce the episode-based training scheme to enhance the adaptability of our model to new tasks. We conduct ablation study to evaluate the effectiveness of our ingenious designs.

\noindent\textbf{Preserving visual feature discriminability}.
% Most existing ZSL approaches choose to project visual features to attribute space or an intermediate space, which however damage the vi
% For the space where we calculate classification score and perform classification, we can choose either the semantic space, an intermediate space (implemented as half of the dimension of the visual space), or the visual feature embedding space. 
To study the importance of preserving visual discriminability, we implement two baseline methods: one we project visual features to attribute space and the other we project visual features to an intermediate space (of half dimension as the visual space). All the other settings are the same as our method. 

Table \ref{Table_ablation} shows that
% the results for both dot product and cosine similarity based score functions in the three respective spaces. 
% We can observe that 
the performance degrades significantly by projecting visual features to either the semantic space or the intermediate space, no matter using 
% as embedding space, the performance is much lower than that using the visual space as the embedding space, for the 
dot product or Cosine similarity based classification score functions.
As analyzed before, image feature embeddings for ZSL are usually generated offline by some powerful feature extraction networks such that high discriminatibility has already been secured. Reprojecting them to either the attribute or the intermediate space shall inevitably impair the discriminability. What is worse, the attribute space or the intermediate space are often of lower dimension than the visual embedding space. The visual variance, which is crucial to ensure discriminability, shall be shrunk once the feature embeddings are reprojected to the lower-dimensional spaces. Due to the damage of the discriminability of visual features, the hubness problem becomes even more intense, leading to much worse results. 

\begin{table}
  \footnotesize
  \renewcommand{\tabcolsep}{4pt}
  \begin{center}
     \begin{tabular}{l|ccccccc}\hline
    % \begin{tabular}{lccccccc}\hline
V$\rightarrow$A                   & \CheckmarkBold      & \CheckmarkBold    &                   &                   &                &                  &   \\ \hline
V $\rightarrow$ I $\leftarrow$ A  &                     &                   & \CheckmarkBold    & \CheckmarkBold    &                &                  &   \\ \hline
A$\rightarrow$V                   &                     &                   &                   &                   & \CheckmarkBold & \CheckmarkBold   & \CheckmarkBold    \\ \hline
Dot product                       & \CheckmarkBold      &                   & \CheckmarkBold    &                   & \CheckmarkBold &                  &                   \\ \hline
Cosine similarity                 &                     & \CheckmarkBold    &                   & \CheckmarkBold    &                &  \CheckmarkBold  &  \CheckmarkBold   \\ \hline
Episode based training            &                     &                   &                   &                   &                &                  &  \CheckmarkBold   \\ \hline \hline
ZSL                               & 36.3              & 45.1              &  34.2               &  42.8                 &   27.0         &  67.7                & 70.9     \\ \hline  
GZSL-U                            & 24.5              & 10.1              &  25.9               &  11.2                 &   22.7         &  59.8                & 62.7     \\ \hline
GZSL-S                            & 62.5              & 86.8              &  68.9               &  81.8                 &   53.2         &  75.2                & 77.0     \\ \hline
GZSL-H                            & 35.2              & 18.0              &  37.6               &  19.6                 &   31.9         &  66.6                & 69.1     \\ \hline
    \end{tabular}
  \end{center}
  \vspace{-5pt}
  % \vspace{-10pt}
  \caption{Ablation study on the AwA1 dataset. ``V$\rightarrow$A'', ``A$\rightarrow$V'', and ``V $\rightarrow$ I $\leftarrow$ A''
  refer to projecting visual features to attribute space, projecting attributes to visual space, and projecting both visual and attribute features into an intermediate space, respectively. }
  \label{Table_ablation}
 % \vspace{-20pt}
 \vspace{-5pt}
\end{table}

% Compared with the embedding space, the classification The embedding space is the dominant factor.

% \subsubsection{Inductive ZSL}  
\noindent\textbf{Cosine similarity based classification score function}.
We compare dot product and cosine similarity based loss functions 
within all the three classification spaces. 
Table \ref{Table_ablation} shows that 
the classification space seems a more dominant factor: neither of the two score functions 
works well if the classification space is not appropriate. When the visual embedding space is selected for classification, the proposed cosine similarity based score function results in much better performance than that based on dot product. 
We speculate the reason is that values of class attribute are not continuous such that there are large variance 
among the attribute vectors of different classes. 
Consequently, classifier weights derived from them also possess large variance,  which might cause high variances of inputs to the Softmax activation function~\cite{luo2017cosine}.
% which results in classifier weights of large variance. 
Unlike dot product, our cosine similarity based score function normalizes the classifier weights before calculating its dot product with visual embeddings. This normalization procedure can bound and reduce the variance of the classifier weights, contributing to better performance.
% resulting in better performances. 
% When calculating cosine similarity, we need normalize the classification weight
\begin{table*}
\footnotesize
  \renewcommand{\tabcolsep}{3.4pt}
  \begin{center}
    \begin{tabular}{l|c|ccc||c|ccc||c|ccc||c|ccc||c|ccc} \hline
& \multicolumn{4}{c|}{\textbf{SUN}} 
& \multicolumn{4}{c|}{\textbf{CUB}}  
& \multicolumn{4}{c|}{\textbf{AWA1}} 
& \multicolumn{4}{c|}{\textbf{AWA2}}
& \multicolumn{4}{c}{\textbf{aPY}} 
\\ \cline{2-21}
& ZSL & \multicolumn{3}{|c|}{GZSL} 
& ZSL & \multicolumn{3}{|c|}{GZSL} 
& ZSL & \multicolumn{3}{|c|}{GZSL} 
& ZSL & \multicolumn{3}{|c|}{GZSL} 
& ZSL & \multicolumn{3}{|c}{GZSL} 
\\ \cline{2-21}
& T1 & U & S & H  
& T1 & U & S & H  
& T1 & U & S & H  
& T1 & U & S & H  
& T1 & U & S & H  
% \\ \hline
% DAP \cite{lampert2014attribute}
% & 39.9 & 4.2 & 25.1 & 7.2 
% % & 40.0 & 1.7 & 67.9 & 3.3 
% & 0.0  & 0.0 & 88.7 & 0.0
% & 46.1 & 0.0 & 84.7 & 0.0 
% & 33.8 & 4.8 & 78.3 & 9.0 
% \\ 
% SSE \cite{zhang2015zero}
% & 51.5 & 2.1 & 36.4 & 4.0 
% % & 43.9 & 8.5 & 46.9 & 14.4 
% & 60.1 & 7.0 & 80.5 & 12.9
% & 61.0 & 8.1 & 82.5 & 14.8 
% & 34.0 & 0.2 & 78.9 & 0.4
% \\ 
% CONSE \cite{norouzi2013zero}       
% & 38.8 & 6.8 & 39.9 & 11.6 
% % & 34.3 & 1.6 & \textbf{72.2} & 3.1
% & 45.6 & 0.4 & 88.6 & 0.8
% & 44.5 & 0.5 & 90.6 & 1.0 
% & 26.9 & 0.0 & \textbf{91.2} & 0.0
% \\ 
% CMT \cite{socher2013zero}
% & 39.9 & 8.1 & 21.8 & 11.8
% % & 34.6 & 7.2 & 49.8 & 12.6
% & 39.5 & 0.9 & 87.6 & 1.8 
% & 37.9 & 0.5 & 90.0 & 1.0 
% & 28.0 & 1.4 & 85.2 & 2.8
\\\hline
LATEM \cite{xian2016latent}
& 55.3  & 14.7  & 28.8  & 19.5 
& 49.3  & 15.2  & 57.3  & 24.0 
& 55.1  & 7.3   & 71.7  & 13.3 
& 55.8  & 11.5  & 77.3  & 20.0 
& 35.2  & 0.1   & 73.0  & 0.2
\\ 
ALE \cite{akata2015evaluation}
& 58.1 & 21.8 & 33.1 & 26.3 
& 54.9 & 23.7 & 62.8 & 34.4  
& 59.9 & 16.8 & 76.1 & 27.5 
& 62.5 & 14.0 & 81.8 & 23.9 
& 39.7 & 4.6  & 73.7 & 8.7
\\
DEVISE \cite{frome2013devise}
& 56.5 & 16.9 & 27.4 & 20.9 
& 52.0 & 23.8 & 53.0 & 32.8 
& 54.2 & 13.4 & 68.7 & 22.4 
& 59.7 & 17.1 & 74.7 & 27.8 
& \textbf{39.8} & 4.9  & 76.9 & 9.2
\\
SJE \cite{akata2015evaluation}
& 53.7 & 14.7 & 30.5 & 19.8 
& 53.9 & 23.5 & 59.2 & 33.6  
& 65.6 & 11.3 & 74.6 & 19.6 
& 61.9 & 8.0  & 73.9 & 14.4 
& 32.9 & 3.7  & 55.7 & 6.9
\\
ESZSL \cite{romera2015embarrassingly}        
& 54.5 & 11.0 & 27.9 & 15.8 
& 53.9 & 12.6 & 63.8 & 21.0 
& 58.2 & 6.6  & 75.6 & 12.1 
& 58.6 & 5.9  & 77.8 & 11.0 
& 38.3 & 2.4  & 70.1 & 4.6
\\
SYNC \cite{changpinyo2016synthesized}         
& 56.3 & 7.9  & \textbf{43.3} & 13.4 
& 55.6 & 11.5 & 70.9 & 19.8 
& 54.0 & 8.9  & 87.3 & 16.2 
& 46.6 & 10.0 & 90.5 & 18.0 
& 23.9 & 7.4  & 66.3 & 13.3 
\\
SAE (\cite{kodirov2017semantic})        
& 40.3  & 8.8  & 18.0  & 11.8 
& 33.3  & 7.8  & 54.0  & 13.6   
& 53.0  & 1.8  & 77.1  & 3.5 
& 54.1  & 1.1  & 82.2  & 2.2 
& 8.3   & 0.4  & 80.9  & 0.9
\\
GFZSL \cite{verma2017simple}
& 60.6 & 0.0 & 39.6 & 0.0 
& 49.3 & 0.0 & 45.7 & 0.0 
& 68.3 & 1.8 & 80.3 & 3.5 
& 63.8 & 2.5 & 80.1 & 4.8 
& 38.4 & 0.0 & 83.3 & 0.0
\\ 
DEM \cite{zhang2017learning}
& 61.9  & 20.5  & 34.3  & 25.6
& 51.7  & 19.6  & 57.9  & 29.2
& 68.4  & 32.8  & 84.7  & 47.3
& 67.2  & 30.5  & 86.4  & 45.1
& 35.0  & 11.1  & 75.1  & 19.4
\\
Relat. Net \cite{yang2018learning}  
& -     & -     & -     & -
& 55.6  & 38.1  & 61.1  & 47.0
& 68.2  & 31.4  & \textbf{91.3}  & 46.7
& 64.2  & 30.0  & \textbf{93.4}  & 45.3
& -     & -     & -     & -
\\
SP-AEN \cite{chen2018zero}
& 59.2  & 24.9  & 38.6  & 30.3
& 55.4  & 34.7  & 70.6  & 46.6
& -     & -     & -     & -
& 58.5  & 23.3  & 90.9  & 37.1
& 24.1  & 13.7  & 63.4  & 22.6
\\
PSR \cite{annadani2018preserving}
& 61.4  & 20.8  & 37.2  & 26.7
& 56.0  & 24.6  & 54.3  & 33.9
& -     & -     & -     & -
& 63.8  & 20.7  & 73.8  & 32.3
& 38.4  & 13.5  & 51.4  & 21.4
\\ \hline
$\textrm{f-CLSWGAN}^{\star}$ \cite{xian2018feature}
& 60.8 & \textbf{42.6} & 36.6 & \textbf{39.4}  
& \textbf{57.3} & \textbf{57.7} & 43.7 & \textbf{49.7}
& 68.2  & 57.9 & 61.4 & 59.6
& - & - & - & -
& - & - & - & - 
% \\
% $\textrm{SE}^{\star}$ \cite{verma2018generalized}
% & 63.4 & 40.9 & 30.5 & 34.9
% & 59.6 & 53.3 & 41.5 & 46.7
% & 69.5 & 56.3 & 67.8 & 61.5
% & 69.2 & 58.3 & 68.1 & 62.8
% & - & - & - & -
% TEDE \cite{zhang2018towards}
% & 62.4 & 22.1 & 35.6 & 27.3 
% & 57.1 & 21.0 & 66.0 & 31.9 
% & 70.1 & 36.9 & 90.6 & 52.4 
% & 66.5 & 35.2 & 93.0 & 51.1 
% & 20.4 & 7.8  & 75.3 & 14.1
% \\
\\ \hline
Ours 
& \textbf{62.6} & 36.3 & 42.8 & 39.3         
& 54.4 & 47.4 & 47.6 & 47.5
& \textbf{70.9} & \textbf{62.7} & 77.0 & \textbf{69.1}    
& \textbf{71.1} & \textbf{56.4} & 81.4 & \textbf{66.7}  
& 38.0 & \textbf{26.5} & 74.0 & \textbf{39.0}
\\
\hline
\end{tabular} 
\end{center}
 \vspace{-8pt}
\caption{Zero-shot learning accuracy. The best results are in \textbf{bold}. The model with $^\star$
(f-CLSWGAN) generates additional data for training while the remaining models do not.}
 \label{result_zsl}
%  \vspace{-10pt}
\end{table*}

\noindent\textbf{Episode-based training mechanism}
The proposed episode-based training mechanism is to train our classifier weight
generator in the way it works during test.
From Table \ref{Table_ablation}, we can observe that there are about 3$\%$ performance gains
for both the ZSL setting and GZSL setting when this unique training mechanism is adopted. 
This is within our expectation because after exposing our weight generator with numerous 
(fake) new ZSL tasks during training, it acquires the knowledge how to deal with real new ZSL tasks during test. So, better performance is more likely to be guaranteed.

\begin{table*}
\footnotesize
  \renewcommand{\tabcolsep}{3.5pt}
  \begin{center}
    \begin{tabular}{l|c|ccc||c|ccc||c|ccc||c|ccc||c|ccc} \hline
& \multicolumn{4}{|c|}{\textbf{SUN}} 
& \multicolumn{4}{|c|}{\textbf{CUB}}  
& \multicolumn{4}{|c|}{\textbf{AWA1}} 
& \multicolumn{4}{|c|}{\textbf{AWA2}}
& \multicolumn{4}{|c}{\textbf{aPY}} 
\\ \cline{2-21}
& ZSL & \multicolumn{3}{|c|}{GZSL} 
& ZSL & \multicolumn{3}{|c|}{GZSL} 
& ZSL & \multicolumn{3}{|c|}{GZSL} 
& ZSL & \multicolumn{3}{|c|}{GZSL} 
& ZSL & \multicolumn{3}{|c}{GZSL} 
\\ \cline{2-21}
& T1 & U & S & H  
& T1 & U & S & H  
& T1 & U & S & H  
& T1 & U & S & H  
& T1 & U & S & H  
\\ \hline
ALE-tran \cite{akata2015evaluation}
& 55.7  & 19.9 & 22.6 & 21.2 
& 54.5  & 23.5 & 45.1 & 30.9
& 65.6  & 25.9 & - & -
& 70.7  & 12.6 & 73.0 & 21.5
& 46.7  & 8.1  & - & -
\\ 
GFZSL-tran \cite{verma2017simple}
& \textbf{64.0} &  0.0 & 41.6 & 0.0
& 49.3 & 24.9 & 45.8 & 32.2
& 81.3 & 48.1 & - & - 
& 78.6 & 31.7 & 67.2 & 43.1
& 37.1 & 0.0 & - & - 
\\ 
DSRL \cite{norouzi2013zero}
& 56.8 &  17.7 & 25.0 & 20.7 
& 48.7 & 17.3 & 39.0 & 24.0  
& 74.7 & 22.3 & - & -
& 72.8 & 20.8 & 74.7 & 32.6
& 45.5 & 11.9 & - & -
\\ 
QFSL \cite{song2018transductive}
& 58.3  & 31.2  & 51.3  & 38.8
& \textbf{72.1}  & \textbf{71.5}  & \textbf{74.9}  & \textbf{73.2}
& -     & -     & -     & -
& 79.7  & 66.2  & \textbf{93.1}  & 77.4
& -     & -     & -     & - 
\\ 
\hline
Ours-trans (XE)
& 61.9 & 44.5 & 57.6 & 50.2
& 59.2 & 54.4 & 67.9 & 60.4
& 87.4 & 84.2 & 84.3 & 84.2
& 81.4 & 77.7 & 88.3 & 82.7
& 52.7 & 50.4 & 86.3 & 63.7
\\
Ours-trans (GXE)
& 63.5  & \textbf{45.4} & \textbf{58.1} & \textbf{51.0}            
& 61.3  & 57.0 & 68.7 & 62.3           
& \textbf{89.8}  & \textbf{87.7} & \textbf{89.0} & \textbf{88.4}
& \textbf{83.2}  & \textbf{80.2} & 90.0 & \textbf{84.8}    
& \textbf{54.7}  & \textbf{51.8} & \textbf{87.6} & \textbf{65.1}            
\\
\hline
    \end{tabular} 
  \end{center}
 \vspace{-5pt}
\caption{Transductive zero-shot learning accuracy. The best results are in \textbf{bold}.}
 \label{result_gzsl}
%  \vspace{-10pt}
\end{table*}

\subsection{Comparative Results}

\noindent\textbf{Zero-shot learning}.
Table \ref{result_zsl} shows the comparative results of the proposed method and the state-of-the-art ones for the inductive ZSL problem. For conventional ZSL, our method reaches the best for three out of the five datasets. Remarkably, for the AwA2 dataset, our method beats the second best by about 4$\%$. 
% It is worthy to note that 
% Remarkably, our method consistently outperforms DEM (\cite{zhang2017learning}) for all the five datasets, which substantiates the benefit of our method of taking consideration of inter-class separability when learning the mapping from semantic space to visual space. 

\noindent\textbf{Generalized zero-shot learning}.
More interesting observations can be made for the GZSL setting where classification is performed over both seen and unseen classes.
% seen classes are also included to be the candidates. 
With more classes involved, the classification accuracy of unseen classes drops for all methods.
However, our method exhibits much more robustness than the other ones and drops moderately on these datasets.  
Remarkably, our method sometimes secures accuracy that is even by about 100$\%$ (\textit{aPY}) higher than the second best. 
% As we will analyze in depth later, % beat the second best 
% Our method significantly outperforms all competing methods, reaching performance gains over the second best even about 30\% in the \textit{AWA1} dataset. We analyze the reason for our dramatic advantage is that 
We analyze this striking improvements are brought by our consideration of inter-class separation during training so that the resultant classifiers for seen classes possess favorable class separation property after training and
% They are highly discriminative to judge if an incoming image belongs to the classes they were trained for.
% When they are combined with classifiers generated from semantic descriptions of unseen classes, 
% the resultant overall classifiers 
% It 
shall be highly discriminative to discern whether an incoming image belongs to the classes they were trained for. 

% From the perspective of hubness problem, since the classifier weights for seen class have good separation property, the weight vectors are less likely to be clustered in the embedding space, so that the risk is reduced that some candidates are selected as the nearest neighbors for many query images. 

Contrary to the striking advantages for recognizing unseen classes, our method seems kind of ``forgetful'' and is overcome by many methods for recognizing seen classes. This is because during training, we constantly sample new ZSL tasks to train the weight generator to acquire the knowledge of handling new ZSL tasks. Unlike existing methods, which process the whole dataset altogether or are specially designed to keep the training memory, our method intentionally forgets the global class structure of training set. Therefore, with the increase of the capability of handle new ZSL tasks, it inevitably sacrifices some competence of classifying seen classes.  Despite of this, our method surpasses the other ones by large margins for three out of the five datasets for the harmonic mean (H), while being very close to the feature synthesized based method, f-CLSWGAN, which generates additional data for training.  
% which measures the overall classification performance for both seen and unseen classes. 
% our method is trained with various 
 % was intended to memorize the global class structure of the whole training set
% memorize the class information of the whole datasets.
% Since different tasks 

\noindent\textbf{Transductive zero-shot learning}.
When test data are available during training, better performance is often expected as we can utilize them to mitigate the classification bias towards seen classes.    
% some specific strategies can be adopted to take advantage of these 
% data and often result in better classifier for the unlabeled images.
% Therefore, better performance is often expected in this setting. 
Table \ref{result_gzsl} verifies this and our transductive algorithm significantly outperforms the inductive counterpart. This substantiates the effectiveness of our novel learning-without-forgetting self-training technique. Further, 
% that compared with the inductive counterpart, 
% the performance of our method is significantly boosted  
% for all the four datasets. Besides, 
with generalized 
cross-entropy loss for unseen classes, Ours-trans (GXE) consistently performs better than 
Ours-trans (XE) which uses conventional cross-entropy loss. This
shows the effectiveness of using the generalized cross-entropy loss 
for avoiding the negative impact of incorrect pseudo labels.
Comparatively speaking, similar as we have observed in the inductive setting,
our method significantly outperforms existing ones, especially for 
unseen classes in GZSL.

% \begin{figure}
%   \begin{center}
%     \includegraphics[width=0.495\linewidth]{./figs/self_training_analysis_sub1.pdf}     
%     \includegraphics[width=0.495\linewidth]{./figs/self_training_analysis_sub2.pdf}     
%   \end{center}
%   \vspace{-10pt} 
% \caption{Analysis of the self-training process on the AwA1 dataset. 
% ``5685'' stands for the total number of test images.}
% \label{training_analysis}   
% \vspace{-10pt} 
% \end{figure}

\begin{figure}
  \begin{center}
    \includegraphics[width=0.495\linewidth]{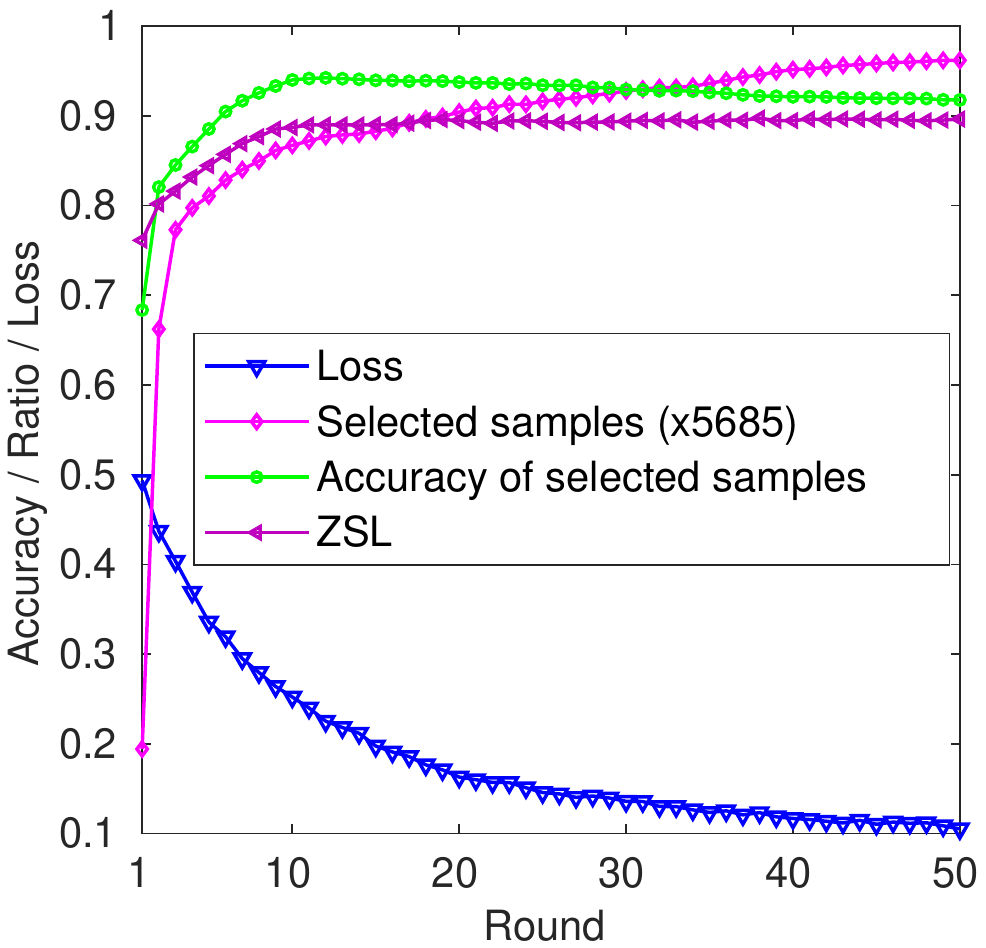}     
    \includegraphics[width=0.495\linewidth]{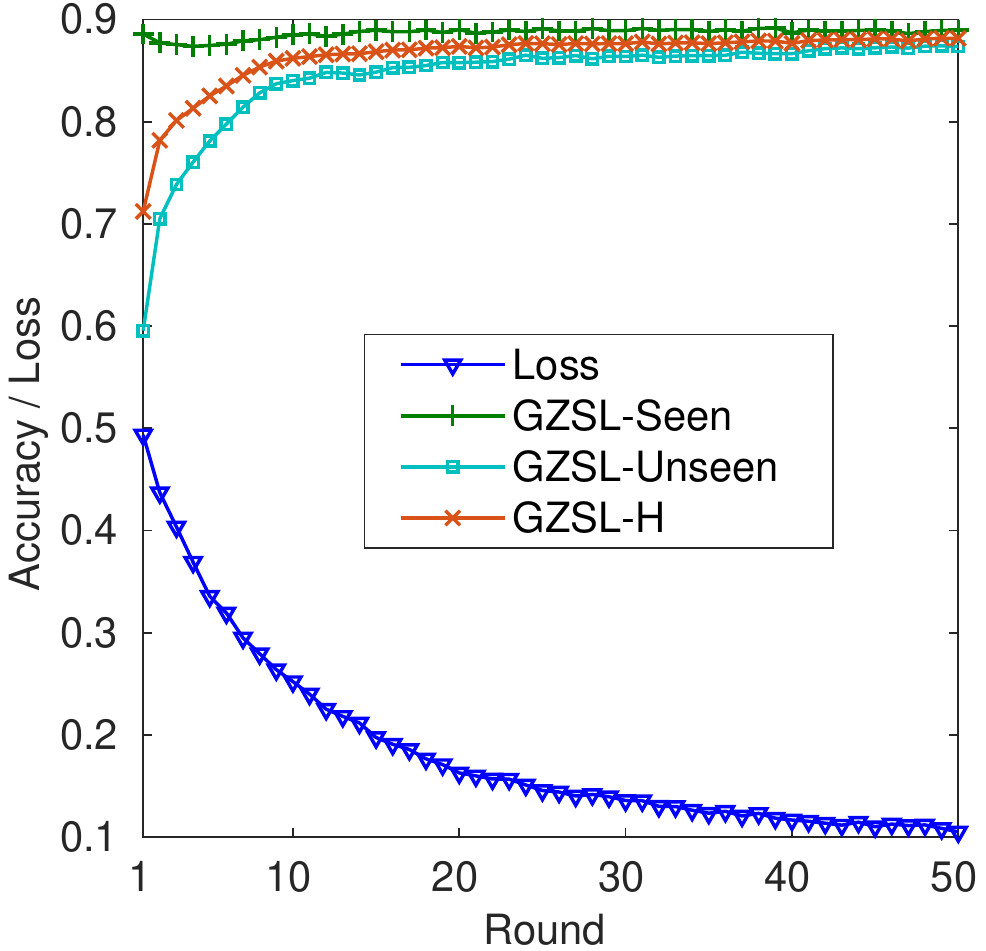}     
  \end{center}
  \vspace{-10pt} 
\caption{Analysis of the self-training process on the AwA1 dataset. 
``5685'' is the total number of test images.}
\label{training_analysis}   
\vspace{-10pt} 
\end{figure}

\begin{figure*}
  \begin{center}
    \includegraphics[width=0.246\linewidth]{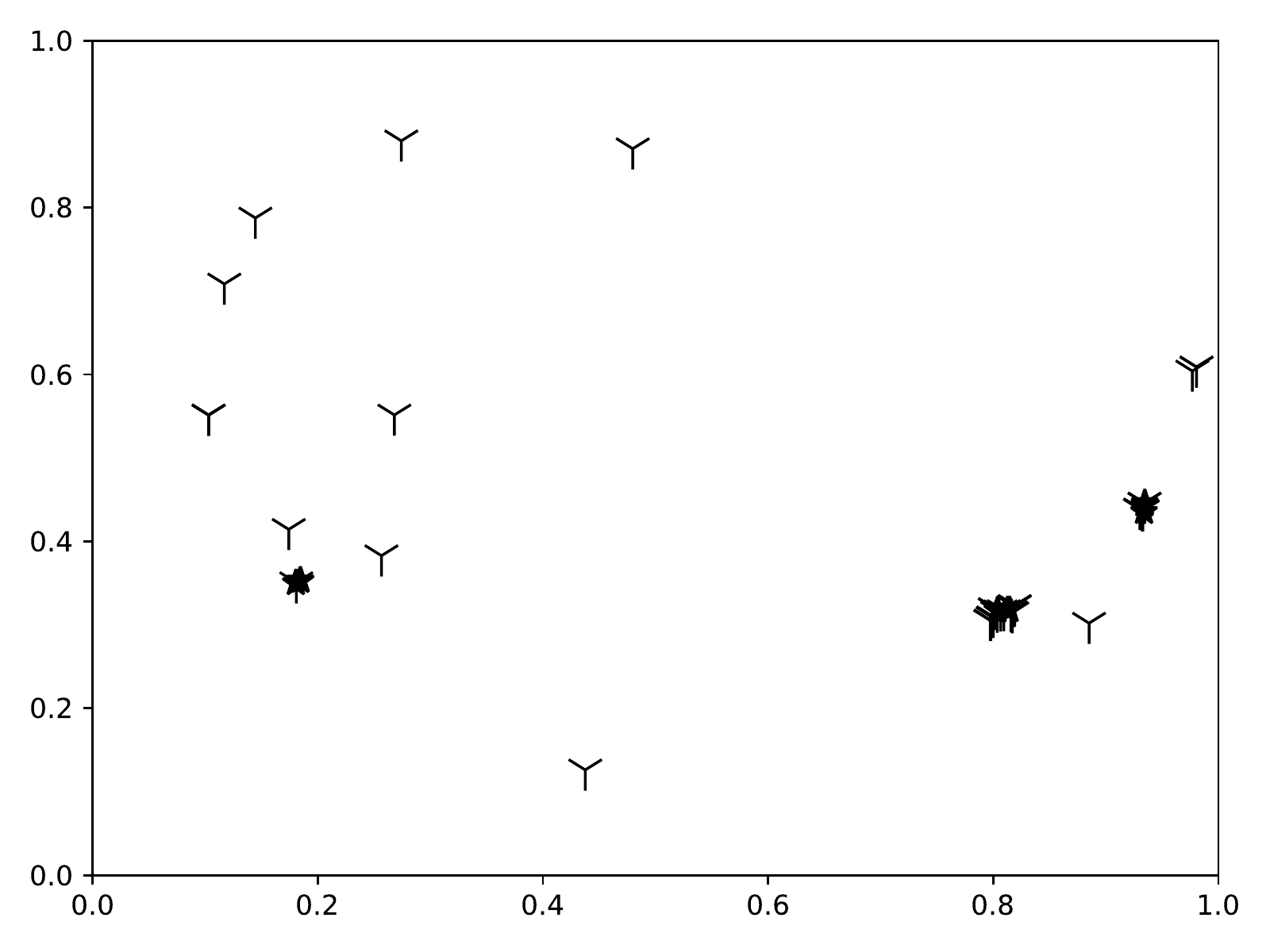}     
    \includegraphics[width=0.246\linewidth]{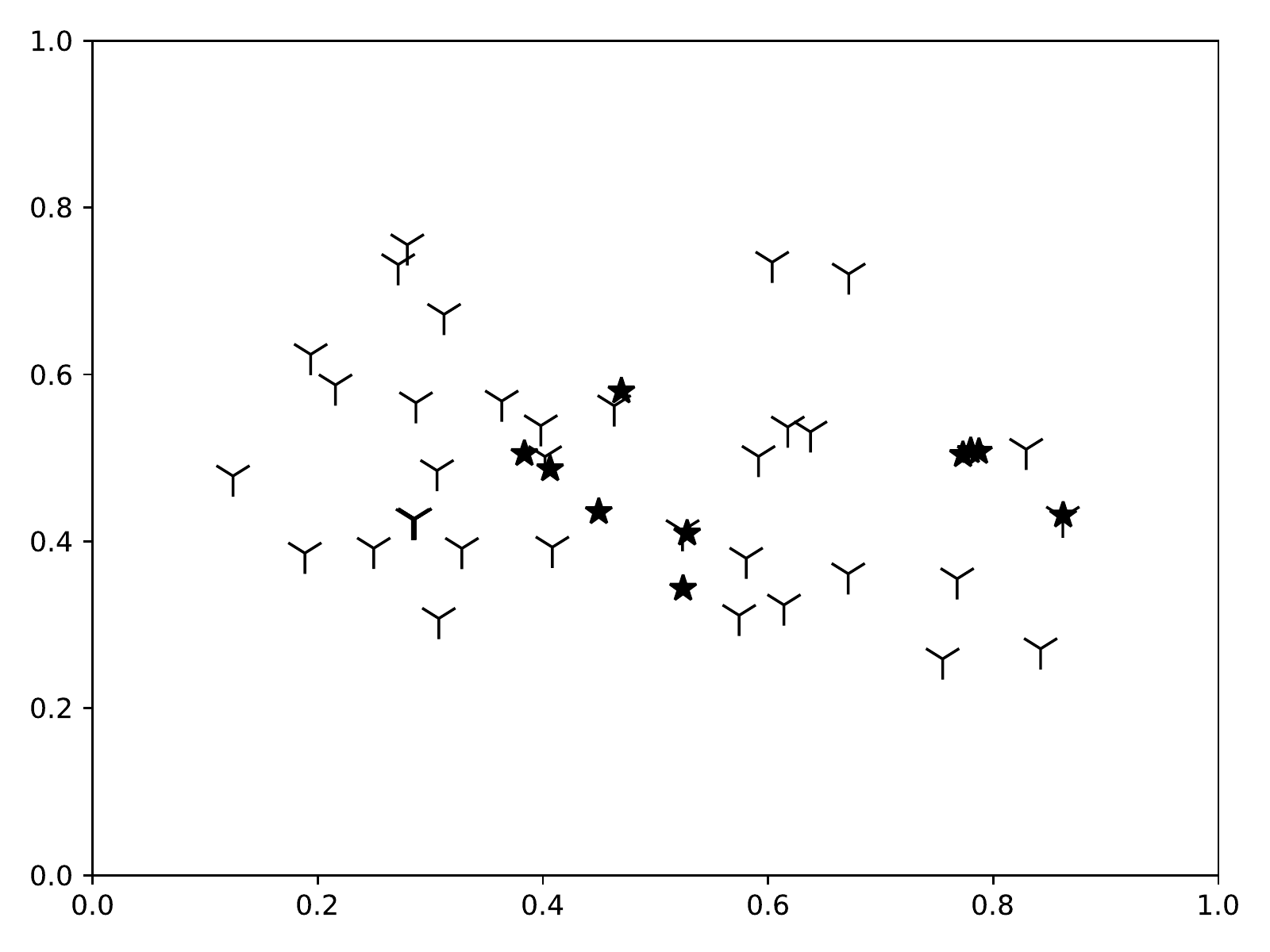}     
    \includegraphics[width=0.246\linewidth]{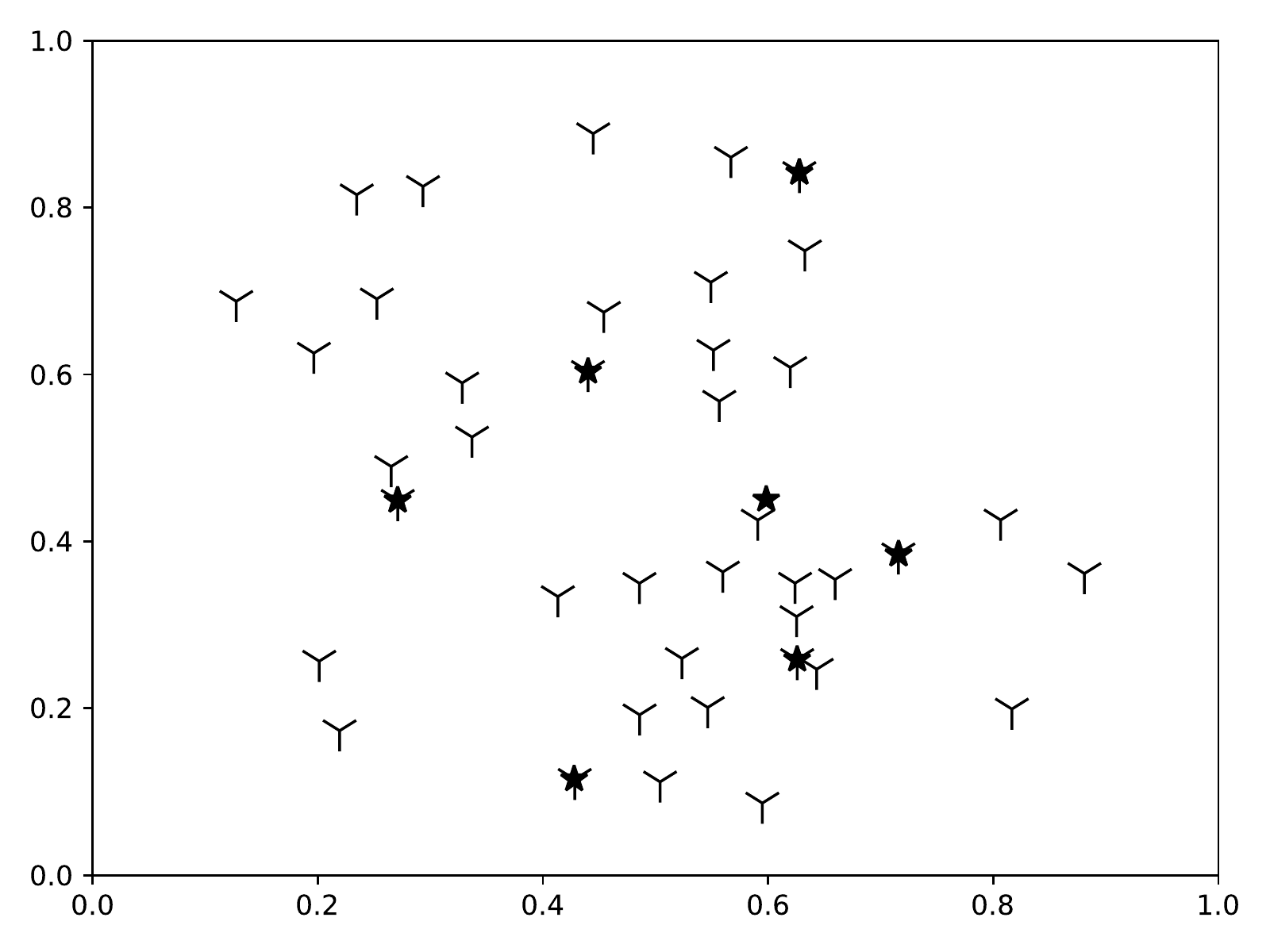}     
    \includegraphics[width=0.246\linewidth]{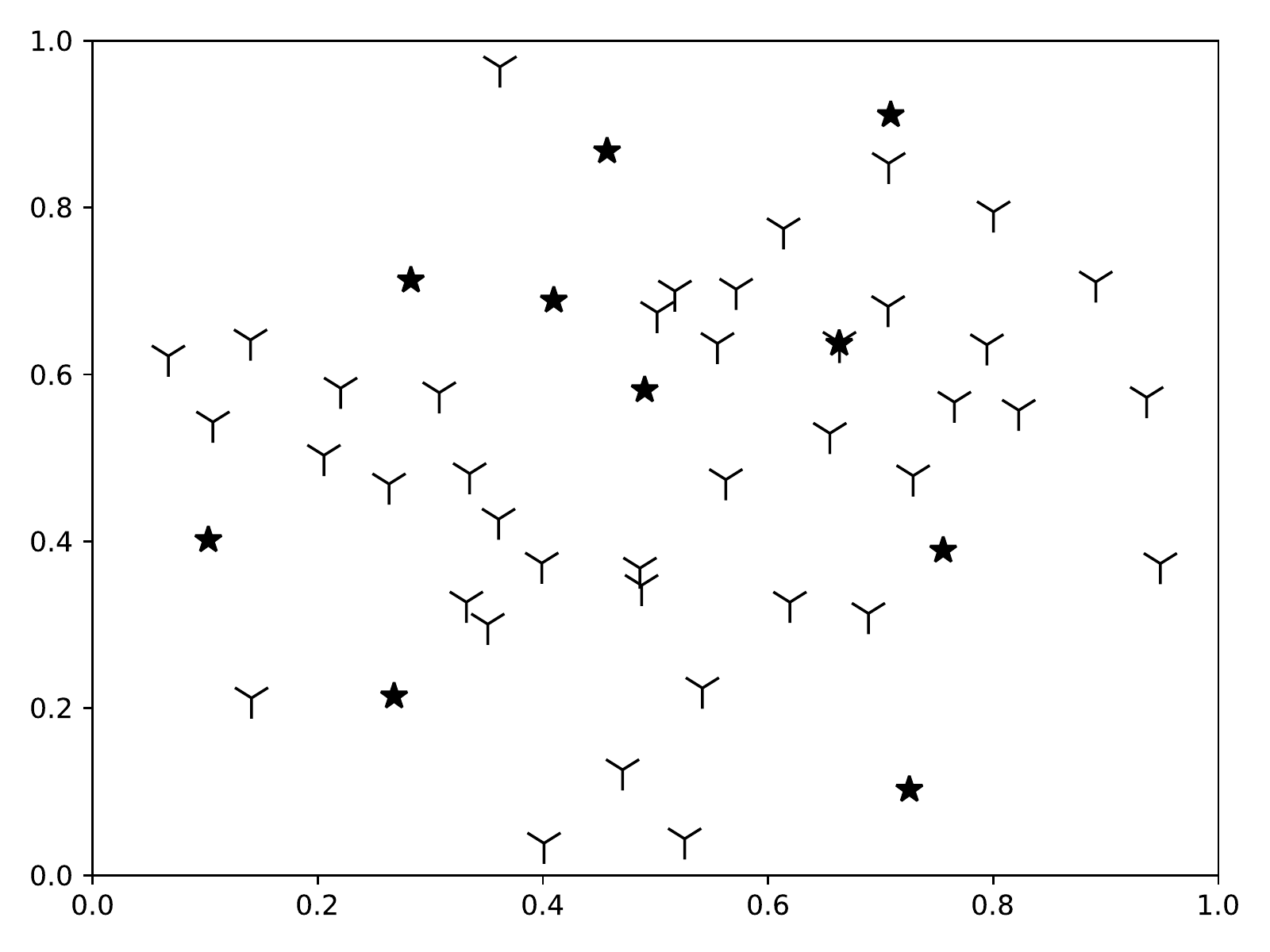}     
    \\
    \includegraphics[width=0.246\linewidth]{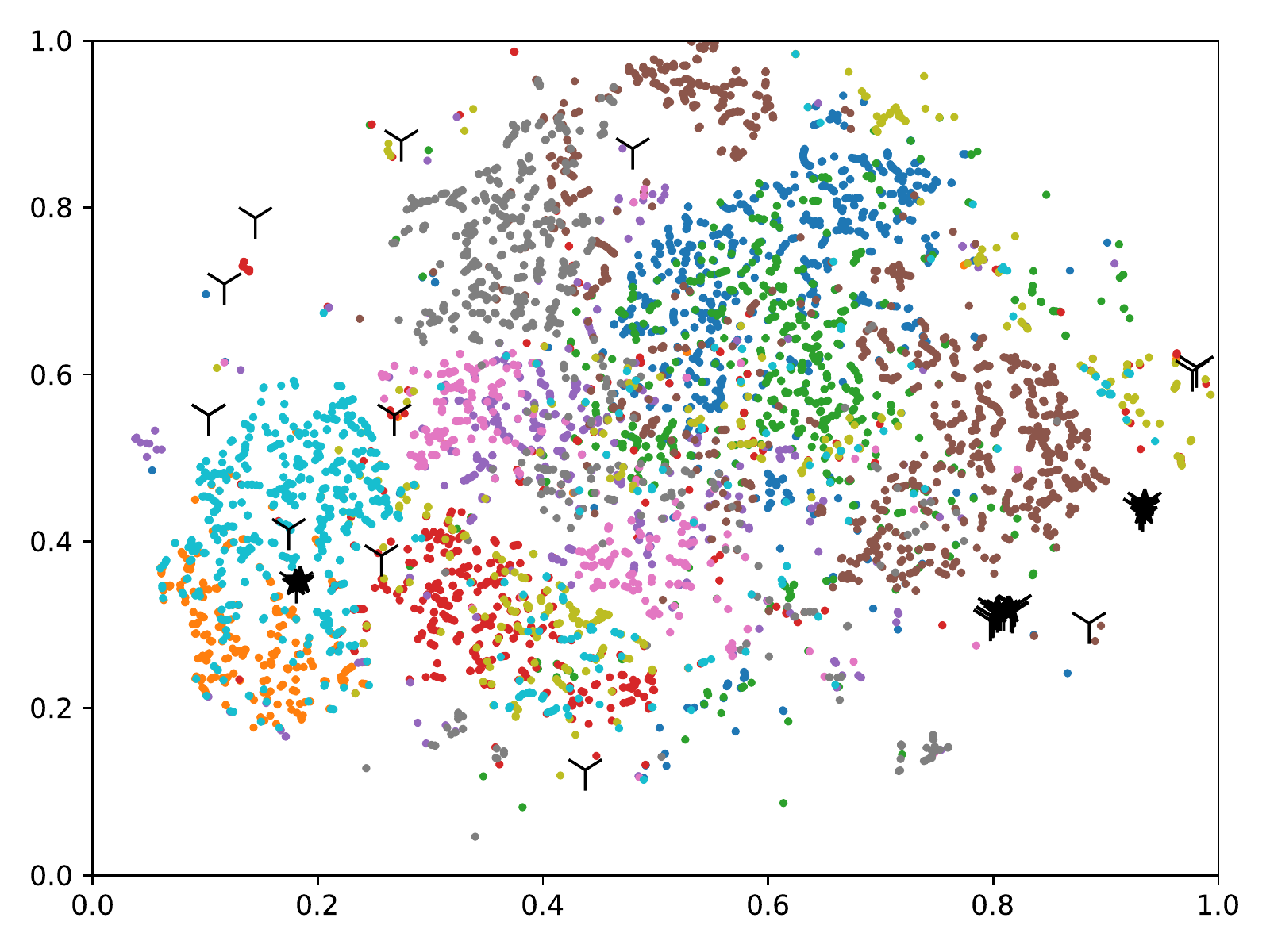}     
    \includegraphics[width=0.246\linewidth]{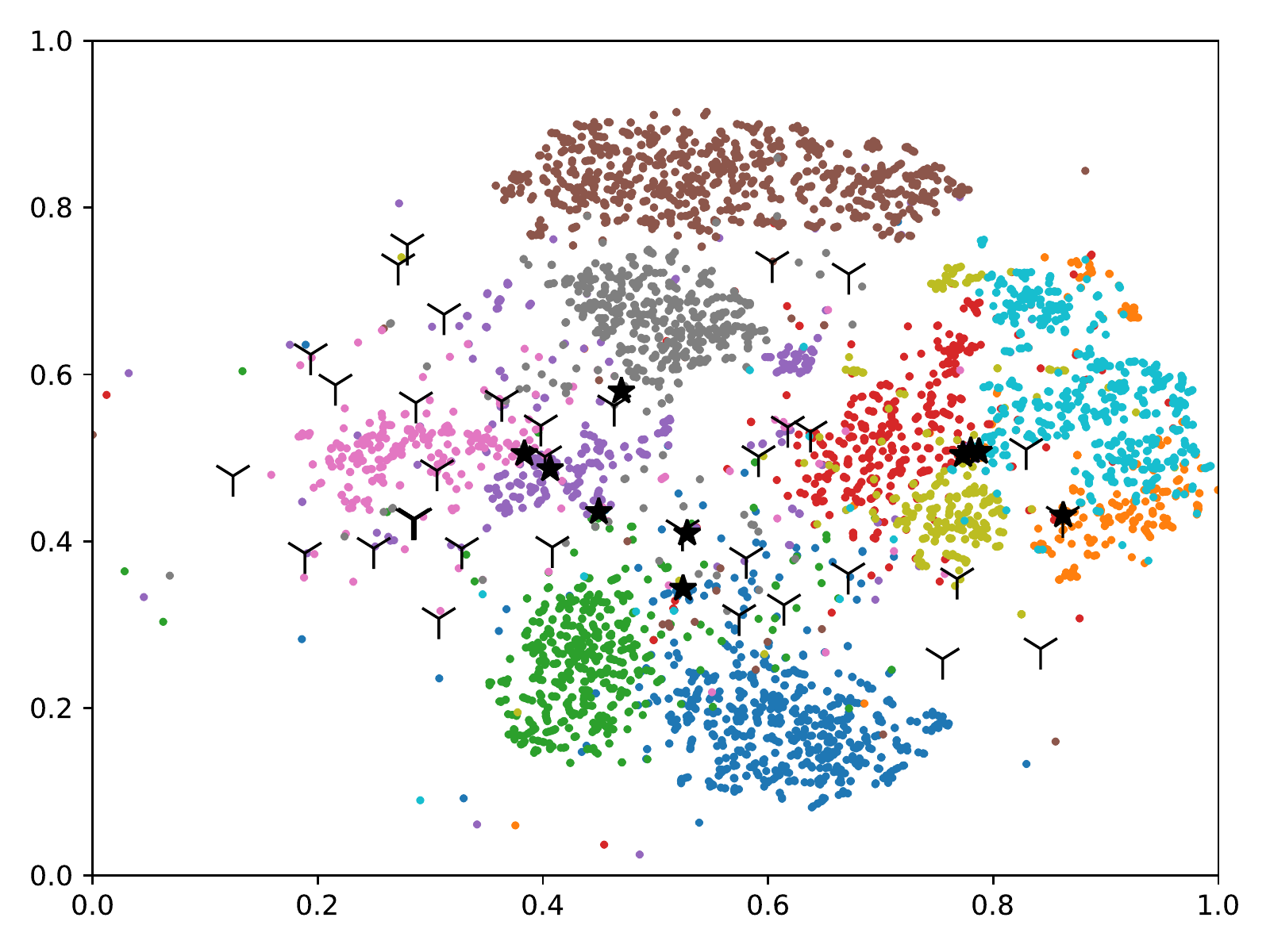}     
    \includegraphics[width=0.246\linewidth]{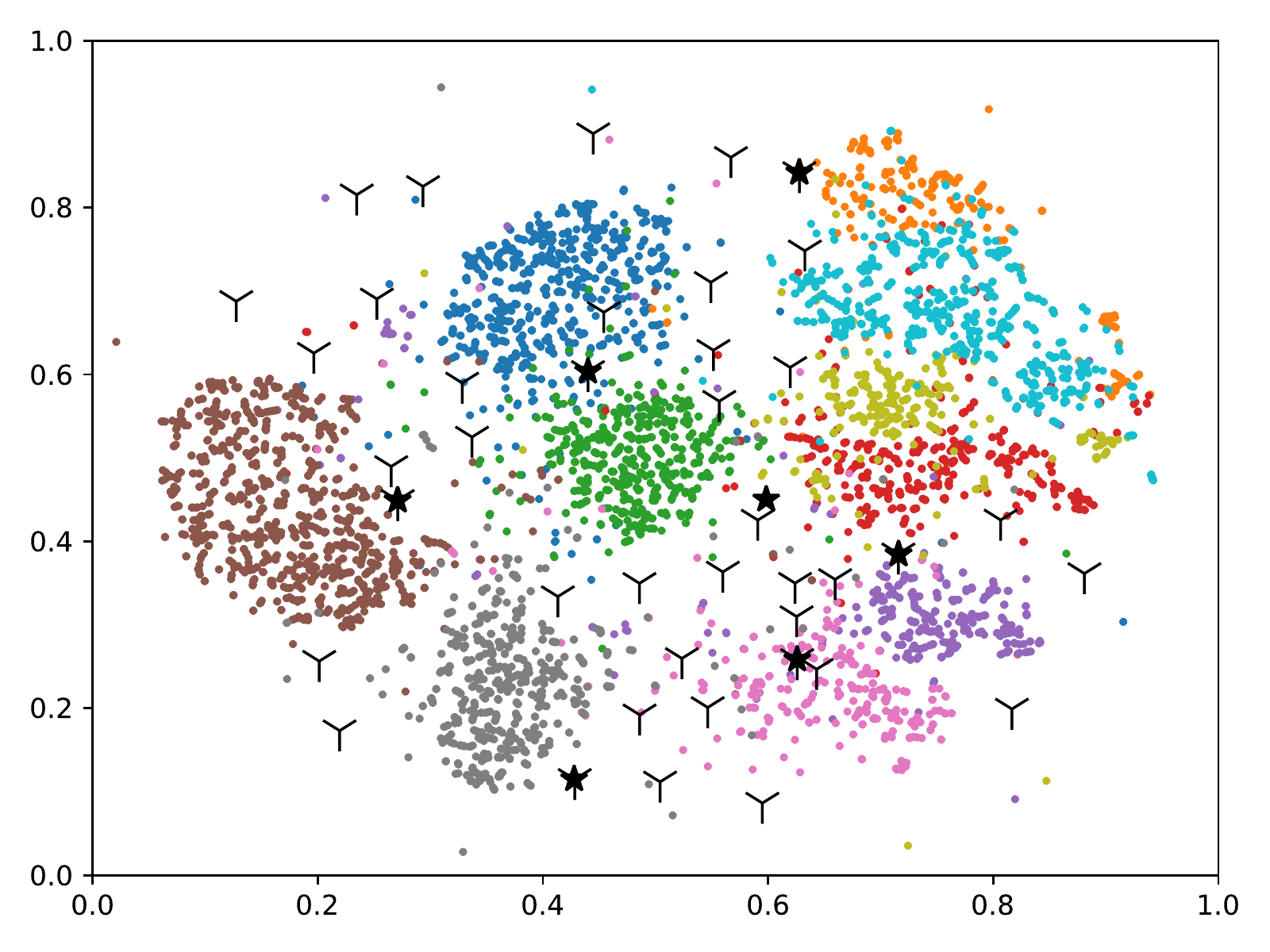}     
    \includegraphics[width=0.246\linewidth]{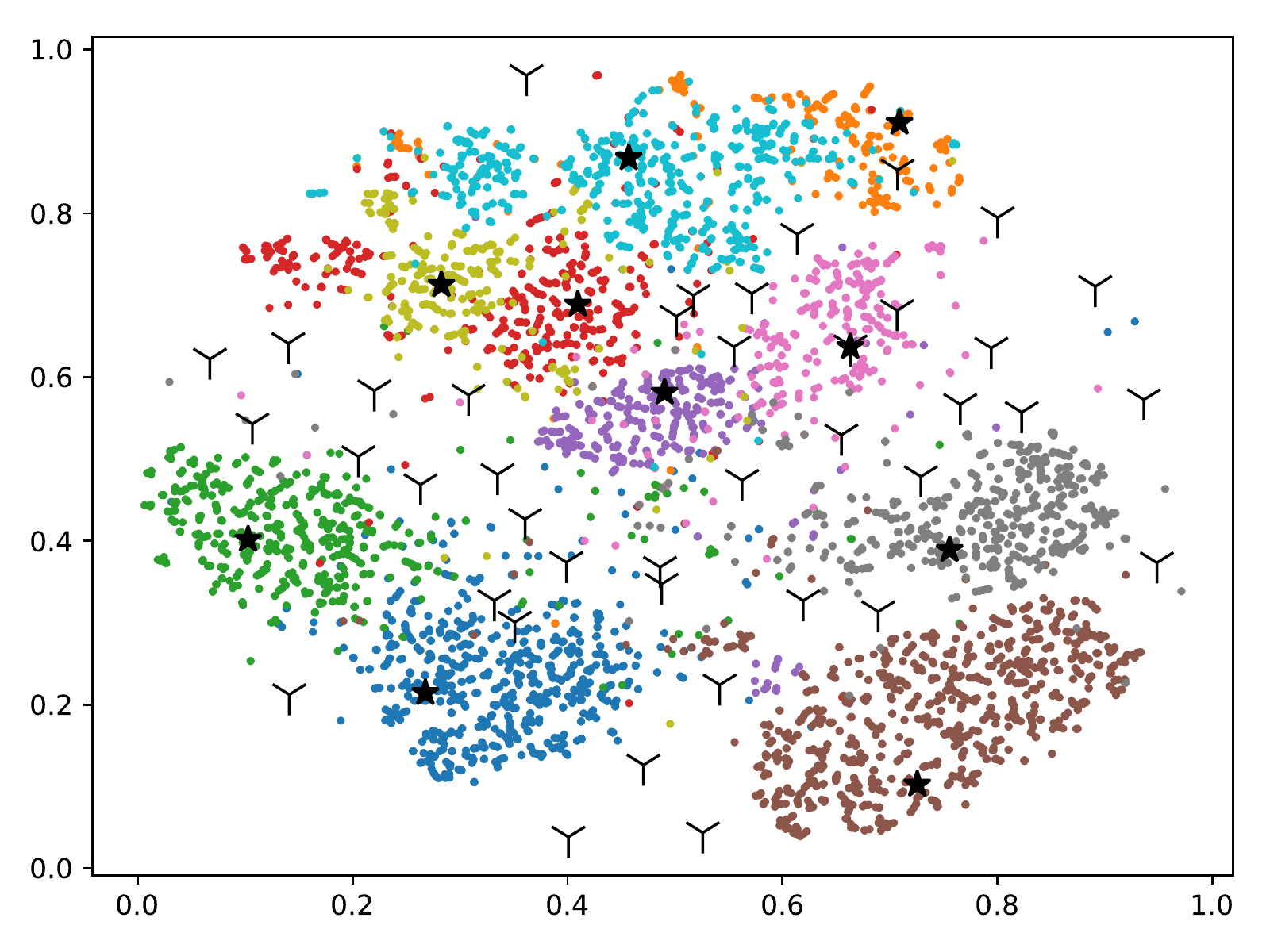}     
    \\
   \hspace{0.2in}  (a) V$\rightarrow$A \hspace{1.1in} (b) DEM \cite{zhang2017learning} \hspace{1.1in} (c) Ours \hspace{1.1in}
    (d) Ours-trans
  \end{center}
  \vspace{-5pt} 
\caption{t-SNE \cite{maaten2008visualizing} visualization of visual feature embeddings and classifier weight vectors (or class prototypes) for the AwA1 dataset. \textbf{Top}: classifier weight vectors (or class prototypes) of both seen (``Y'') and unseen ($\star$) classes. \textbf{Bottom}: classifier weight vectors and visual feature embeddings for unseen classes. Different colors represent different classes. 
% \textbf{Bottom}:  classifier weight vectors and visual feature embeddings for seen classes. 
``V$\rightarrow$A'' represents projecting visual embeddings to attribute space.
}
\label{tsne}  
\vspace{-5pt} 
\end{figure*}

\subsection{Further Analyses}
% \noindent\textbf{Effectiveness of self-training}
\noindent\textbf{Analyzing self-training process}.
In the transductive ZSL setting, we propose to calibrate weight generator $f$ towards unseen classes using test data in a novel self-training fashion. 
We alternate between generating pseudo labels for unseen images using $f$ and updating $f$ using the pseudo labels of high confidence. By this self-training strategy, the bias of 
$f$ towards seen classes can be progressively eliminated, with boost for unseen class recognition as the consequence. 

To analyze how this self-training process works,
% demonstrate this, 
we plot in Figure \ref{training_analysis} the changes of training loss, classification accuracy, number of confident unseen samples (used for updating the model) and the portion of the correctly labeled ones among them.
We can see that as the training round increases, the training loss keeps decreasing and the collection of confident samples is consistently enlarged. At the same time, the accuracy of pseudo label assignment is also promoted. This means with the increase of training round, the unlabeled images used for training are boosted in terms of both quantity and quality, which in return further improves the classifier generator. 
% contributes to a even better w
 % Consequently, all the classification accuracy evaluation metrics are significantly promoted. 

\noindent\textbf{Number of classes per episode}.
Table \ref{tab:zs_cls} shows that ZSL accuracy changes little w.r.t. sampled classes in each mini-batch, which contradicts the observations in \cite{snell2017prototypical}, where episode-based training is used for few-shot learning. 
We speculate the reason is that sampling more classes per mini-batch in \cite{snell2017prototypical} helps boost discriminability of the feature extraction model, as it is required to extract distinct features for more classes in each mini-batch.
This does not apply to us as we use pretrained features. Sampling more classes in each mini-batch to train the classifier generator can be approximated by sampling multiple mini-batches.

\begin{table}
   \footnotesize
     \renewcommand{\tabcolsep}{10.pt}
   \begin{center}
   
      \begin{tabular}{|l|c|c|c|c|c|} \hline 
& c=4 & c=8 & c=16 & c=32 & c=40 (all)
\\\hline
ZSL 
& 68.1
& 69.6
& 70.4
& 70.9 
& 69.8
\\ 
\hline
\end{tabular}
\end{center} 
\vspace{-10pt} 
\caption{ZSL accuracy w.r.t. training classes per batch.}
\label{tab:zs_cls}
\vspace{-10pt} 
\end{table}

\noindent\textbf{Embedding visualization}.
Recall that we calculate the possibility of an image $\textbf{x}$ of belonging to class $y$ given class attribute $\textbf{a}_y$ by calculating the Cosine similarity of $\textbf{x}$ and the classifier weight $\textbf{w}_y$ generating from $\textbf{a}_y$ (Eq. \eqref{probability}). As Cosine similarity of two vectors is equivalent to their dot product after being normalized,
we can view $\frac{\textbf{w}_y}{\|\textbf{w}_y\|}$ as the prototype of class $y$. By this interpretation, the possibility of $\textbf{x}$ of belonging to class $y$ can be measured by the distance of
the normalized feature $\frac{\textbf{x}}{\|\textbf{x}\|}$ and the normalized classifier weight vector $\frac{\textbf{w}_y}{\|\textbf{w}_y\|}$.
Thus, we can visualize normalized classifier weight vectors and normalized visual feture vectors to 
 % and analyze their spatial distribution with respect to visual features to 
 qualitatively evaluate the discriminability of the classifiers. 

We plot the t-SNE visualizations \cite{maaten2008visualizing} of the classifier weights
and their overlappings with the visual features of unseen classes in Figure \ref{tsne}. We can see that our class prototypes are more spatially dispersed than that of DEM \cite{zhang2017learning} which does not consider the inter-class separation information for generating class prototypes. Besides, we can observe that by projecting visual features to attribute space, the corresponding class prototypes are extremely clustered. This substantiates the merits of formulating ZSL as a conditional visual classification problem, by which we can naturally benefit from the high discrimination of the visual features and the inter-class separation information to get discriminative classifiers for both seen and unseen classes. Moreover, we can also see that the distribution of the class prototypes in the transductive setting is even more dispersed than that for the inductive setting. This evidences the effectiveness of our transductive ZSL algorithm in exploiting unlabeled test data for enhancing the discriminability of the classifiers for both seen and unseen classes. 

By overlapping the class prototypes with visual features of unseen classes, we can observe that visual features of unseen classes lie closely with their corresponding class prototypes, whiling being far away from those of seen classes.  
% the prototypes of seen classes lie closely with their corresponding visual features of seen classes, while being far from those of unseen classes, and vice versa. 
In contrast, this favorable distribution cannot be observed in the plots of DEM and the algorithm which projects visual features to the attribute space. This further substantiates the superiority of our method.

% we also visualize classification weight of our method after 
% good separation property

% generated from the attributes of seen classes show good separation property so that the hubness problem is not as severe as that for other methods. 
% The hubness problem refers that in ZSL, some candidate points are prone to be the nearest neighbors of many query points when the dimension is high. So, if the candidate points are more evenly distributed in the space, the less severe of the hubness problem should be. 
% To validate this, we use t-SNE \cite{maaten2008visualizing} to visualize the classification weight vectors generated from all 50 class semantic vectors in the AWA1 dataset. As a comparison, we do the same thing for DEM (\cite{zhang2017learning}) which also learns mapping from semantic space to visual space. The result is shown in Figure \ref{tsne}. We can observe that the points are more evenly distributed for our method than that for DEM. This further validates the benefit of our method in avoiding the hubness problem.

\section{Conclusions}
In this paper, we reformuate ZSL as a visual feature classification problem conditioned on the attributes. %In this new framework, we can naturally benefit from the high discriminability of visual feature and the inter-class separation information for classification.
Under this reformulation, we develop algorithms for various ZSL settings. For the conventional setting, we propose to learn a deep neural network to generate visual feature classifiers directly from the attributes, and guide the process with a cosine similarity based cross-entropy loss and an episode-based training scheme. For the generalized setting, we propose to concatenate classifiers for both seen and unseen classes to recognize objects from all classes. For the transductive setting, we develop a novel learning-without-forgetting self-training mechanism to calibrate the classifier genereator towards unseen classes while maintaining good performance for seen classes. 
% We extend our method % revisit the neural network generation technique for Zero-Shot Learning (ZSL) and reveal that it can be more effectively utilized by selecting visual feature space as the classification space, using cosine similarity based classification score function, and employing episode based training. 
% We also extend the proposed method to the genelized ZSL and transductive ZSL setting by concanating classifiers from both seen and unseen classes.
 % 
Experiments on widely used datasets verify the effectiveness of the proposed methods and demonstrate that the proposed methods obtain remarkable advantages over the state-of-the-art methods, especially for unseen classes in the generalized ZSL setting. 
% The embedding visualization results reveal that the advantages of our methods are due to the good separability of the class weights of seen classes. 
% In the future, we plan to extend our framework to the setting of few-shot learning.

\section*{Appendix}

\section{Further Analysis on the GZSL Performance}
From the experiments in the main article, we can observe that the proposed method reaches significant performance gains over the existing ones for the generalized ZSL setting. Here, we give more explanations for our impressive performance.

\begin{figure}
  \begin{center}
    \includegraphics[width=1.0\linewidth]{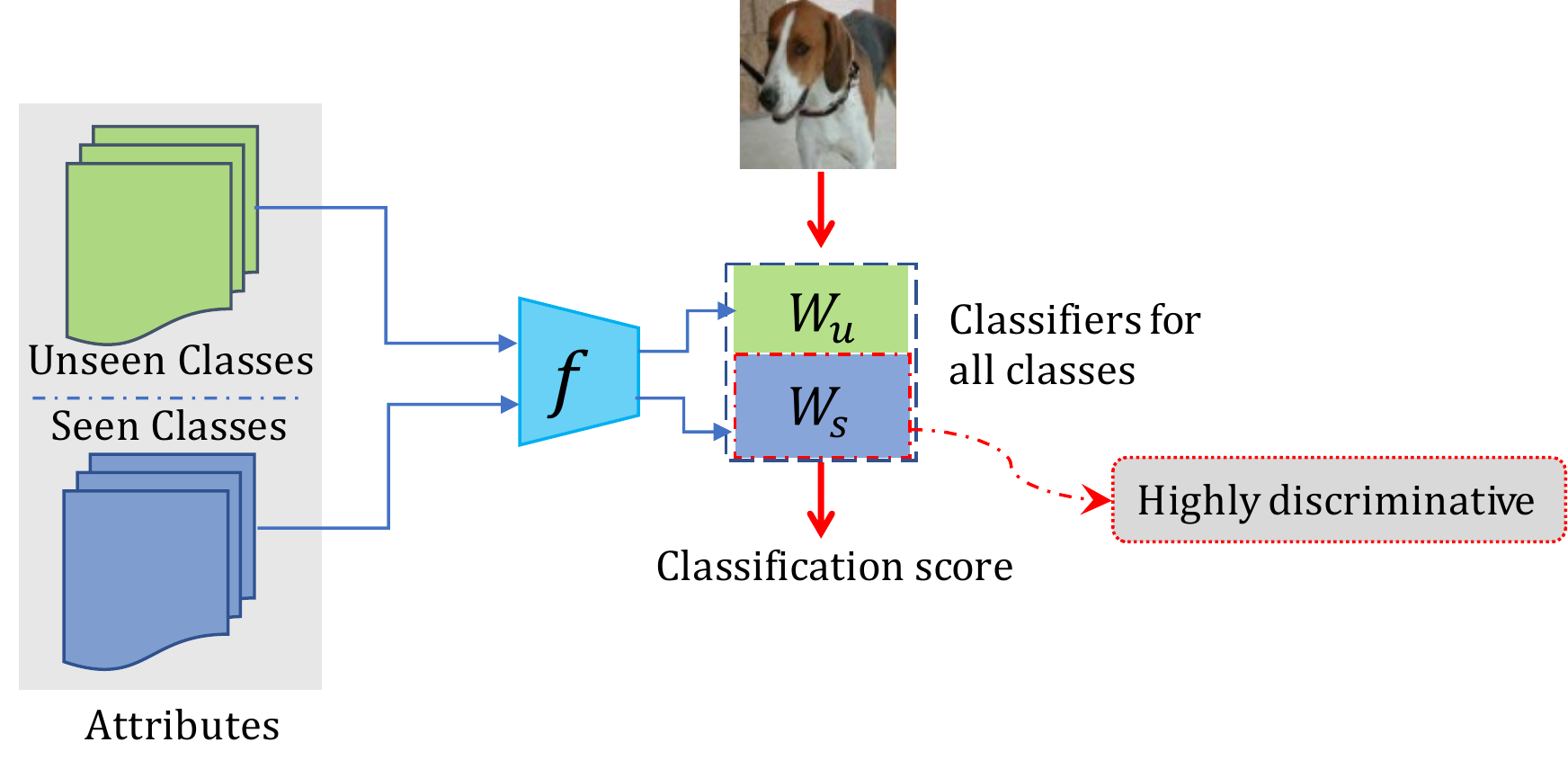}         
  \end{center} 
  \vspace{-15pt}
\caption{Illustration of how we conduct test in the generalized ZSL setting. $\textbf{W}_u$ and $\textbf{W}_s$ are the classifier weight matrices for unseen and seen classes, respectively. We concatenate $\textbf{W}_u$ and $\textbf{W}_s$ to classify images from all classes. Since $f$ is trained with seen classes during training, $\textbf{W}_s$ should be highly discriminative to discern whether an incoming image belongs to seen or unseen classes.
}
\label{generalized_zsl}   
%\vspace{-15pt}
\end{figure}

As explained in the main text, our great advantages in GZSL is owing to our novel problem formulation of ZSL as a conditional visual classification problem.
Due to this formulation, during the test stage, we generate the classifiers for both seen and unseen classes from the corresponding attributes,
and combine (by concatenating the classifier weight matrices) them to classify images from all classes. Figure \ref{generalized_zsl} illustrates the process. \textit{Since our classifier generation model is trained with seen classes, during test, the classifiers for seen classes generated from the corresponding attributes should be highly discriminative to discern whether or not an incoming image belongs to the classes observed during training.} 
Thus, the involvement of seen classes during GZSL test impacts much less on our method than on the existing ones, leading to our much better recognition results.
 
%This explains why our method reaches significant 
%The resulting a classification model for all classes.  %we perform GZSL

\begin{figure}
  \begin{center}
    \includegraphics[width=0.9\linewidth]{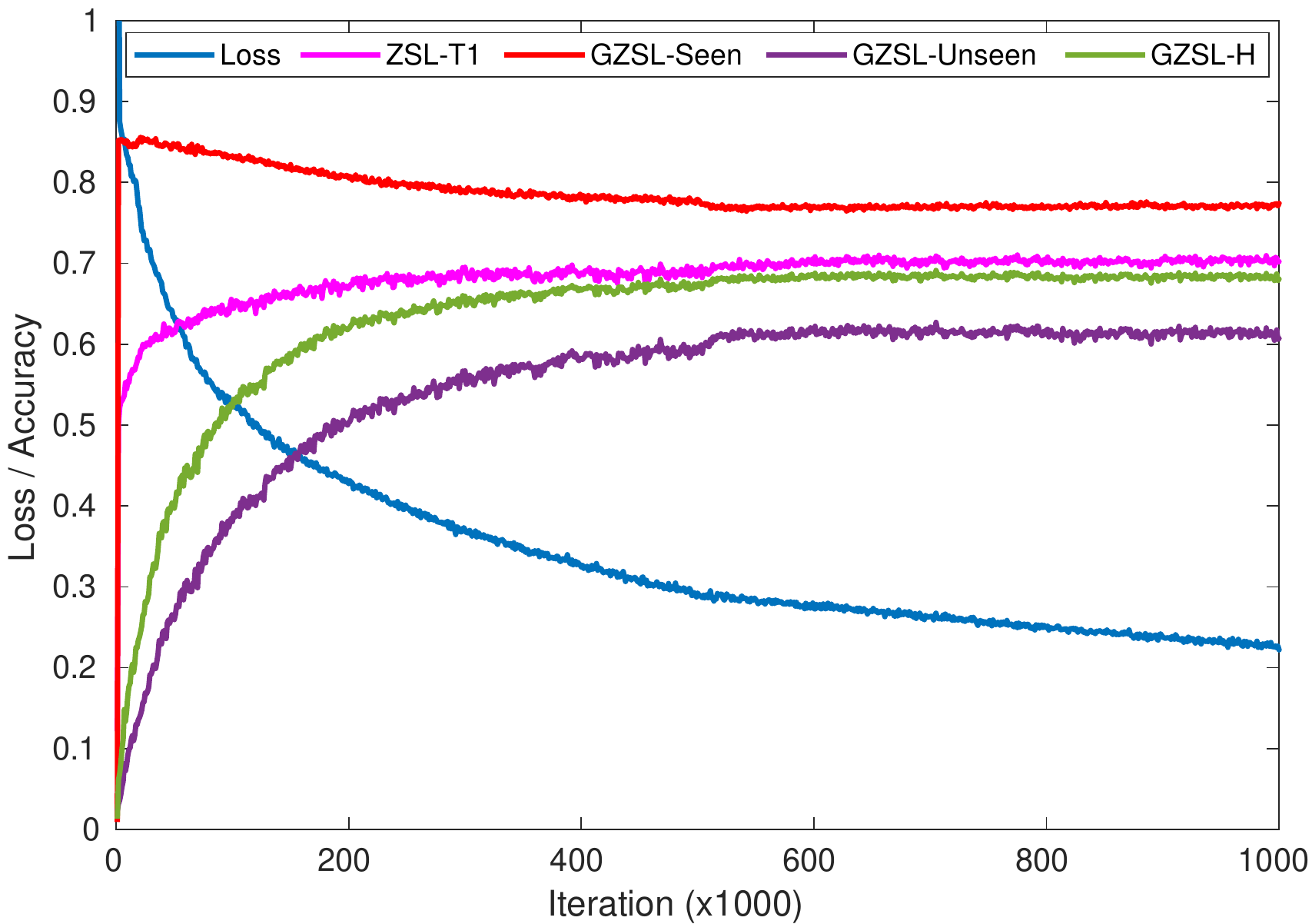}         
  \end{center} 
  \vspace{-15pt}
\caption{Changes of loss and accuracy with training iteration. As the loss decreases with training iteration, the ZSL accuracy (ZSL-T1) keeps increasing and then remains stable. Meanwhile, in the generalized setting, the accuracy for unseen class recognition (GZSL-Unseen) keeps increasing and then remains stable as well, but the accuracy for seen classes (GZSL-Seen) quickly reaches its peak, then deceases, and eventually remains unchanged. Since the decreasing rate of GZSL-Seen is much smaller than increasing rate of GZSL-Unseen, their harmonic mean (GZSL-H) has the same changing pace as GZSL-Unseen, \textit{i.e.}, increasing first and then remaining stable.
}
\label{change_of_accuracy}   
%\vspace{-15pt}
\end{figure}

Another thing to be further noted is that in the GZSL experimental setting, our performance for seen classes is less competitive and is often inferior to the state-of-the-art. To figure out how this happens, we plot in Figure~\ref{change_of_accuracy} the changes of performance on the AwA1 dataset with respect to the training iteration. We can observe that in the conventional ZSL setting, the accuracy (ZSL-T1) first keeps increasing and then remains stable, along with the decrease of the training loss. For the generalized ZSL setting, the accuracy for unseen classes (GZSL-Unseen) and seen classes (GZSL-Seen) has quite different changing trajectories: GZSL-Seen reaches the peak in the very beginning, drops thereafter, and remains stable later, while GZSL-Unseen first keeps increasing and then remains stable. The dropping rate of GZSL-Seen is much slower than the increasing rate of GZSL-Unseen, which makes their harmonic mean GZSL-H change similarly as GZSL-Unseen.

The plot indicates that our classifier generator acquires quickly the knowledge of classifying the observed classes and reaches the peak performance for seen class recognition. It is then tuned to be apt for categorizing unseen classes as exposed with various randomly sampled new ZSL tasks. As a side effect of the drift towards recognizing unseen classes, the classification boundaries for seen classes turn vaguer, but still remain a fair discriminative level. The enhancement in the competence of recognizing unseen classes, in combined with a fair maintenance of the capability of recognizing seen classes, leads to our distinguished performance in GZSL.

\section{Classification Result Visualizations}
\begin{figure*}
  \begin{center}
    \includegraphics[width=1.0\linewidth]{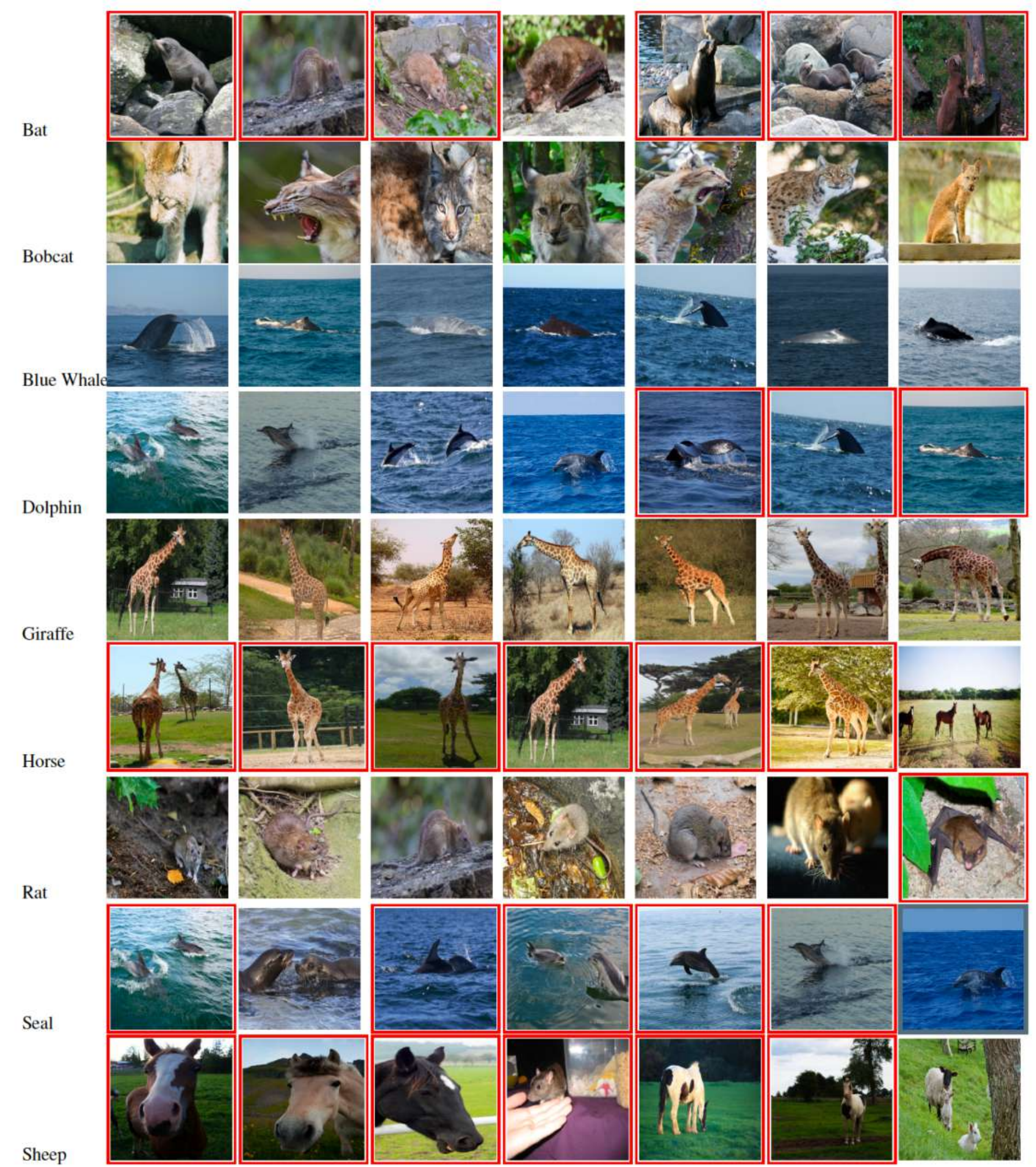}         
  \end{center} 
  \vspace{-15pt}
\caption{Top-7 predictions for the unseen classes based on classification score on \textit{\textbf{AwA2}}. Misclassifications are in red border.}
% \label{change_of_accuracy}   
%\vspace{-15pt}
\label{zsl_awa2}     
\end{figure*}

To facilitate analysis, we visualize the classification results of our method in the conventional ZSL setting.  Figure \ref{zsl_awa2} shows the visualizations on the \textit{AWA1}. In the figure, according to the classification score, we show the top image returns of a class given the semantic description of the class. 

According to the top images, we can see that our method reasonably captures discriminative visual properties of each unseen class based solely on its semantic embedding. We can also see that the misclassified images are with appearance so similar to that of predicted class that even humans cannot easily distinguish between the two.  For example, the ``bat'' images in the first row of Figure \ref{zsl_awa2} look so similar to that of the ``rat'' images. Without carefully observation, human can sometimes make mistakes in differentiating them, even that we have seen various images about the two classes before. Considering that the attributes of the two classes are very similar and our model has never ``seen'' any images of the two classes, it is reasonable to make the mistakes.

% \begin{figure}
% \small
% \begin{center}
% \includegraphics[width=1.0\linewidth]{./figs/merge_crop.pdf}
% \end{center}
% \vspace{-10pt}
% \caption{Top-7 predictions for the unseen classes based on classification score on \textit{\textbf{AwA2}}. Misclassifications are in red border.}
% \label{zsl_awa2}     
% \end{figure}

\section*{Acknowledgement}
This research is supported in part by the NSF IIS award 1651902, U.S. Army Research Office Award W911NF-17-1-0367, and NEC labs America.

% \clearpage

{\small
\bibliographystyle{ieee_fullname}
\bibliography{egbib}

\begin{thebibliography}{10}\itemsep=-1pt

\bibitem{akata2015evaluation}
Zeynep Akata, Scott Reed, Daniel Walter, Honglak Lee, and Bernt Schiele.
\newblock Evaluation of output embeddings for fine-grained image
  classification.
\newblock In {\em CVPR}, 2015.

\bibitem{annadani2018preserving}
Yashas Annadani and Soma Biswas.
\newblock Preserving semantic relations for zero-shot learning.
\newblock In {\em CVPR}, 2018.

\bibitem{changpinyo2016synthesized}
Soravit Changpinyo, Wei-Lun Chao, Boqing Gong, and Fei Sha.
\newblock Synthesized classifiers for zero-shot learning.
\newblock In {\em CVPR}, 2016.

\bibitem{changpinyo2017predicting}
Soravit Changpinyo, Wei-Lun Chao, and Fei Sha.
\newblock Predicting visual exemplars of unseen classes for zero-shot learning.
\newblock In {\em ICCV}, 2017.

\bibitem{chen2018zero}
Long Chen, Hanwang Zhang, Jun Xiao, Wei Liu, and Shih-Fu Chang.
\newblock Zero-shot visual recognition using semantics-preserving adversarial
  embedding network.
\newblock In {\em CVPR}, 2018.

\bibitem{yizhe_zsl_2017}
Mohamed Elhoseiny, Yizhe Zhu, Han Zhang, and Ahmed Elgammal.
\newblock Link the head to the “beak”: Zero shot learning from noisy text
  description at part precision.
\newblock In {\em CVPR}, 2017.

\bibitem{farhadi2009describing}
Ali Farhadi, Ian Endres, Derek Hoiem, and David Forsyth.
\newblock Describing objects by their attributes.
\newblock In {\em CVPR}, 2009.

\bibitem{felix2018multi}
Rafael Felix, BG~Vijay Kumar, Ian Reid, and Gustavo Carneiro.
\newblock Multi-modal cycle-consistent generalized zero-shot learning.
\newblock In {\em ECCV}, 2018.

\bibitem{finn2017model}
Chelsea Finn, Pieter Abbeel, and Sergey Levine.
\newblock Model-agnostic meta-learning for fast adaptation of deep networks.
\newblock {\em arXiv preprint arXiv:1703.03400}, 2017.

\bibitem{frome2013devise}
Andrea Frome, Greg~S Corrado, Jon Shlens, Samy Bengio, Jeff Dean, Tomas
  Mikolov, et~al.
\newblock Devise: A deep visual-semantic embedding model.
\newblock In {\em NIPS}, 2013.

\bibitem{gidaris2018dynamic}
Spyros Gidaris and Nikos Komodakis.
\newblock Dynamic few-shot visual learning without forgetting.
\newblock In {\em CVPR}, 2018.

\bibitem{he2016deep}
Kaiming He, Xiangyu Zhang, Shaoqing Ren, and Jian Sun.
\newblock Deep residual learning for image recognition.
\newblock In {\em CVPR}, 2016.

\bibitem{kodirov2015unsupervised}
Elyor Kodirov, Tao Xiang, Zhenyong Fu, and Shaogang Gong.
\newblock Unsupervised domain adaptation for zero-shot learning.
\newblock In {\em ICCV}, 2015.

\bibitem{kodirov2017semantic}
Elyor Kodirov, Tao Xiang, and Shaogang Gong.
\newblock Semantic autoencoder for zero-shot learning.
\newblock In {\em CVPR}, 2017.

\bibitem{lampert2014attribute}
Christoph~H Lampert, Hannes Nickisch, and Stefan Harmeling.
\newblock Attribute-based classification for zero-shot visual object
  categorization.
\newblock {\em IEEE Transactions on Pattern Analysis and Machine Intelligence},
  36(3):453--465, 2014.

\bibitem{lei2015predicting}
Jimmy Lei~Ba, Kevin Swersky, Sanja Fidler, et~al.
\newblock Predicting deep zero-shot convolutional neural networks using textual
  descriptions.
\newblock In {\em CVPR}, 2015.

\bibitem{li2018support}
Kai Li, Zhengming Ding, Kunpeng Li, Yulun Zhang, and Yun Fu.
\newblock Support neighbor loss for person re-identification.
\newblock In {\em ACM MM}, 2018.

\bibitem{li2019on}
Kai Li, Martin~Renqiang Min, Bing Bai, Yun Fu, and Hans Peter~Graf.
\newblock On novel object recognition: A unified framework for discriminability
  and adaptability.
\newblock In {\em CIKM}, 2019.

\bibitem{li2019gain}
Kunpeng Li, Ziyan Wu, Kuan-Chuan Peng, Jan Ernst, and Yun Fu.
\newblock Guided attention inference network.
\newblock {\em IEEE Transactions on Pattern Analysis and Machine Intelligence
  (TPAMI)}, 2019.

\bibitem{li2019attnbn}
Kunpeng Li, Yulun Zhang, Kai Li, Yuanyuan Li, and Yun Fu.
\newblock Attention bridging network for knowledge transfer.
\newblock In {\em ICCV}, 2019.

\bibitem{li2019vsr}
Kunpeng Li, Yulun Zhang, Kai Li, Yuanyuan Li, and Yun Fu.
\newblock Visual semantic reasoning for image-text matching.
\newblock In {\em ICCV}, 2019.

\bibitem{li2018discriminative}
Yan Li, Junge Zhang, Jianguo Zhang, and Kaiqi Huang.
\newblock Discriminative learning of latent features for zero-shot recognition.
\newblock In {\em CVPR}, 2018.

\bibitem{luo2017cosine}
Chunjie Luo, Jianfeng Zhan, Lei Wang, and Qiang Yang.
\newblock Cosine normalization: Using cosine similarity instead of dot product
  in neural networks.
\newblock {\em arXiv preprint arXiv:1702.05870}, 2017.

\bibitem{maaten2008visualizing}
Laurens van~der Maaten and Geoffrey Hinton.
\newblock Visualizing data using t-sne.
\newblock {\em Journal of machine learning research}, 9(Nov):2579--2605, 2008.

\bibitem{mishra2018generative}
Ashish Mishra, Shiva Krishna~Reddy, Anurag Mittal, and Hema~A Murthy.
\newblock A generative model for zero shot learning using conditional
  variational autoencoders.
\newblock In {\em CVPR Workshops}, 2018.

\bibitem{morgado2017semantically}
Pedro Morgado and Nuno Vasconcelos.
\newblock Semantically consistent regularization for zero-shot recognition.
\newblock In {\em CVPR}, 2017.

\bibitem{norouzi2013zero}
Mohammad Norouzi, Tomas Mikolov, Samy Bengio, Yoram Singer, Jonathon Shlens,
  Andrea Frome, Greg~S Corrado, and Jeffrey Dean.
\newblock Zero-shot learning by convex combination of semantic embeddings.
\newblock In {\em ICLR}, 2014.

\bibitem{patterson2012sun}
Genevieve Patterson and James Hays.
\newblock Sun attribute database: Discovering, annotating, and recognizing
  scene attributes.
\newblock In {\em CVPR}, 2012.

\bibitem{rohrbach2013transfer}
Marcus Rohrbach, Sandra Ebert, and Bernt Schiele.
\newblock Transfer learning in a transductive setting.
\newblock In {\em NIPS}, 2013.

\bibitem{romera2015embarrassingly}
Bernardino Romera-Paredes and Philip Torr.
\newblock An embarrassingly simple approach to zero-shot learning.
\newblock In {\em ICML}, 2015.

\bibitem{schwartz2018delta}
Eli Schwartz, Leonid Karlinsky, Joseph Shtok, Sivan Harary, Mattias Marder,
  Rogerio Feris, Abhishek Kumar, Raja Giryes, and Alex~M Bronstein.
\newblock Delta-encoder: an effective sample synthesis method for few-shot
  object recognition.
\newblock {\em arXiv preprint arXiv:1806.04734}, 2018.

\bibitem{snell2017prototypical}
Jake Snell, Kevin Swersky, and Richard Zemel.
\newblock Prototypical networks for few-shot learning.
\newblock In {\em NIPS}, 2017.

\bibitem{song2018selective}
Jie Song, Chengchao Shen, Jie Lei, An-Xiang Zeng, Kairi Ou, Dacheng Tao, and
  Mingli Song.
\newblock Selective zero-shot classification with augmented attributes.
\newblock In {\em ECCV}, 2018.

\bibitem{song2018transductive}
Jie Song, Chengchao Shen, Yezhou Yang, Yang Liu, and Mingli Song.
\newblock Transductive unbiased embedding for zero-shot learning.
\newblock In {\em CVPR}, 2018.

\bibitem{yang2018learning}
Flood Sung, Yongxin Yang, Li Zhang, Tao Xiang, Philip~HS Torr, and Timothy~M
  Hospedales.
\newblock Learning to compare: Relation network for few-shot learning.
\newblock In {\em CVPR}, 2018.

\bibitem{verma2017simple}
Vinay~Kumar Verma and Piyush Rai.
\newblock A simple exponential family framework for zero-shot learning.
\newblock In {\em ECML-PKDD}, 2017.

\bibitem{vinyals2016matching}
Oriol Vinyals, Charles Blundell, Tim Lillicrap, Daan Wierstra, et~al.
\newblock Matching networks for one shot learning.
\newblock In {\em NIPS}, 2016.

\bibitem{wang2018zero}
Xiaolong Wang, Yufei Ye, and Abhinav Gupta.
\newblock Zero-shot recognition via semantic embeddings and knowledge graphs.
\newblock In {\em CVPR}, 2018.

\bibitem{welinder2010caltech}
Peter Welinder, Steve Branson, Takeshi Mita, Catherine Wah, Florian Schroff,
  Serge Belongie, and Pietro Perona.
\newblock Caltech-ucsd birds 200.
\newblock 2010.

\bibitem{xian2016latent}
Yongqin Xian, Zeynep Akata, Gaurav Sharma, Quynh Nguyen, Matthias Hein, and
  Bernt Schiele.
\newblock Latent embeddings for zero-shot classification.
\newblock In {\em CVPR}, 2016.

\bibitem{xian2018zero}
Yongqin Xian, Christoph~H Lampert, Bernt Schiele, and Zeynep Akata.
\newblock Zero-shot learning-a comprehensive evaluation of the good, the bad
  and the ugly.
\newblock {\em IEEE Transactions on Pattern Analysis and Machine Intelligence},
  2018.

\bibitem{xian2018feature}
Yongqin Xian, Tobias Lorenz, Bernt Schiele, and Zeynep Akata.
\newblock Feature generating networks for zero-shot learning.
\newblock In {\em CVPR}, 2018.

\bibitem{zhang2017learning}
Li Zhang, Tao Xiang, and Shaogang Gong.
\newblock Learning a deep embedding model for zero-shot learning.
\newblock In {\em CVPR}, 2017.

\bibitem{zhang2019mst}
Yulun Zhang, Chen Fang, Yilin Wang, Zhaowen Wang, Zhe Lin, Yun Fu, and Jimei
  Yang.
\newblock Multimodal style transfer via graph cuts.
\newblock In {\em ICCV}, 2019.

\bibitem{zhang2019rnan}
Yulun Zhang, Kunpeng Li, Kai Li, Bineng Zhong, and Yun Fu.
\newblock Residual non-local attention networks for image restoration.
\newblock In {\em ICLR}, 2019.

\bibitem{zhang2018generalized}
Zhilu Zhang and Mert~R Sabuncu.
\newblock Generalized cross entropy loss for training deep neural networks with
  noisy labels.
\newblock {\em arXiv preprint arXiv:1805.07836}, 2018.

\bibitem{zhang2015zero}
Ziming Zhang and Venkatesh Saligrama.
\newblock Zero-shot learning via semantic similarity embedding.
\newblock In {\em ICCV}, 2015.

\bibitem{zhu2018generative}
Yizhe Zhu, Mohamed Elhoseiny, Bingchen Liu, Xi Peng, and Ahmed Elgammal.
\newblock A generative adversarial approach for zero-shot learning from noisy
  texts.
\newblock In {\em CVPR}, 2018.

\bibitem{yizhe_zsl_2018}
Yizhe Zhu, Mohamed Elhoseiny, Bingchen Liu, Xi Peng, and Ahmed Elgammal.
\newblock A generative adversarial approach for zero-shot learning from noisy
  texts.
\newblock In {\em CVPR}, 2018.

\bibitem{yizhe_abp_2019}
Yizhe Zhu, Jianwen Xie, Bingchen Liu, and Ahmed Elgammal.
\newblock Learning feature-to-feature translator by alternating
  back-propagation for generative zero-shot learning.
\newblock In {\em ICCV}, 2019.

\end{thebibliography}
}

\end{document}